\documentclass[10pt,journal,compsoc]{IEEEtran}
\ifCLASSOPTIONcompsoc
  \usepackage[nocompress]{cite}
\else
  \usepackage{cite}
\fi
\ifCLASSINFOpdf
\else
\fi
\usepackage{amsmath}
\usepackage{array}

\ifCLASSOPTIONcompsoc
 \usepackage[caption=false,font=footnotesize,labelfont=sf,textfont=sf]{subfig}
\else
 \usepackage[caption=false,font=footnotesize]{subfig}
\fi
\usepackage{fixltx2e}

\usepackage{stfloats}
\usepackage{url}
\usepackage{subfiles}
\usepackage{cleveref}
\usepackage{graphicx}
\usepackage{amssymb}
\usepackage{booktabs}
\usepackage{mathtools}
\usepackage{multirow}
\usepackage{xcolor}
\usepackage{tablefootnote}
\newcommand{\red}[1]{#1}

\creflabelformat{equation}{#2\textup{#1}#3}
\Crefname{equation}{Eq.}{Eqs.}
\Crefname{figure}{Fig.}{Figs.}
\Crefname{tabular}{Tab.}{Tabs.}

\begin{document}
%
\title{Ultra Fast Deep Lane Detection with Hybrid Anchor Driven Ordinal Classification}
%
%
%
%

\author{Zequn~Qin,
        Pengyi~Zhang,
        and~Xi~Li,~\IEEEmembership{Senior Member,~IEEE}
\IEEEcompsocitemizethanks{\IEEEcompsocthanksitem Z. Qin and P. Zhang are with the College of Computer Science and Technology, Zhejiang University, Hangzhou, Zhejiang, China, 310007. \protect\\
\IEEEcompsocthanksitem X. Li is with the College of Computer Science and Technology, Zhejiang University, Hangzhou, China, 310007; Shanghai Institute for Advanced Study, Zhejiang University, Shanghai, China, 201203; and the Zhejiang–Singapore Innovation and AI Joint Research Lab.

E-mail: zequnqin@gmail.com, pyzhang@zju.edu.cn, xilizju@zju.edu.cn

}
\thanks{(Corresponding author: Xi Li)}}

%
%

\markboth{IEEE TRANSACTIONS ON PATTERN ANALYSIS AND MACHINE INTELLIGENCE}%
{Shell \MakeLowercase{\textit{et al.}}: Bare Demo of IEEEtran.cls for Computer Society Journals}
%



\IEEEtitleabstractindextext{%
\begin{abstract}
Modern methods mainly regard lane detection as a problem of pixel-wise segmentation, which is struggling to address the problems of efficiency and challenging scenarios like severe occlusions and extreme lighting conditions. Inspired by human perception, the recognition of lanes under severe occlusions and extreme lighting conditions is mainly based on contextual and global information. Motivated by this observation, we propose a novel, simple, yet effective formulation aiming at ultra fast speed and the problem of challenging scenarios. Specifically, we treat the process of lane detection as an anchor-driven ordinal classification problem using global features. First, we represent lanes with sparse coordinates on a series of hybrid (row and column) anchors. With the help of the anchor-driven representation, we then reformulate the lane detection task as an ordinal classification problem to get the coordinates of lanes. Our method could significantly reduce the computational cost with the anchor-driven representation. Using the large receptive field property of the ordinal classification formulation, we could also handle challenging scenarios. Extensive experiments on four lane detection datasets show that our method could achieve state-of-the-art performance in terms of both speed and accuracy. A lightweight version could even achieve 300+ frames per second(FPS). Our code is at \underline{\url{https://github.com/cfzd/Ultra-Fast-Lane-Detection-v2}}.
\end{abstract}

}

\maketitle

\IEEEdisplaynontitleabstractindextext

%
\IEEEpeerreviewmaketitle

\IEEEraisesectionheading{\section{Introduction}\label{sec_introduction}}
\IEEEPARstart{L}{ane} detection is a fundamental component in autonomous driving and advanced driver-assistance systems (ADAS) \cite{hillel2014recent}, which distinguishes and locates the lane markings on the road. Although great success has been achieved with the deep learning models, there are still some important and challenging problems to be addressed. 

The first one is the efficiency problem. In practice, the lane detection algorithm is heavily executed to provide instant perception results with constrained vehicle computing devices for the downstream tasks, which require fast detection speed. 
Moreover, previous lane detection methods \cite{SCNN, SAD, Lee_2017_ICCV,End-to-End} are mainly based on segmentation, which is formulated in a dense bottom-up learning pipeline making it hard to achieve fast speed.

\begin{figure}[t]
    \centering
    \includegraphics[width=0.95\linewidth]{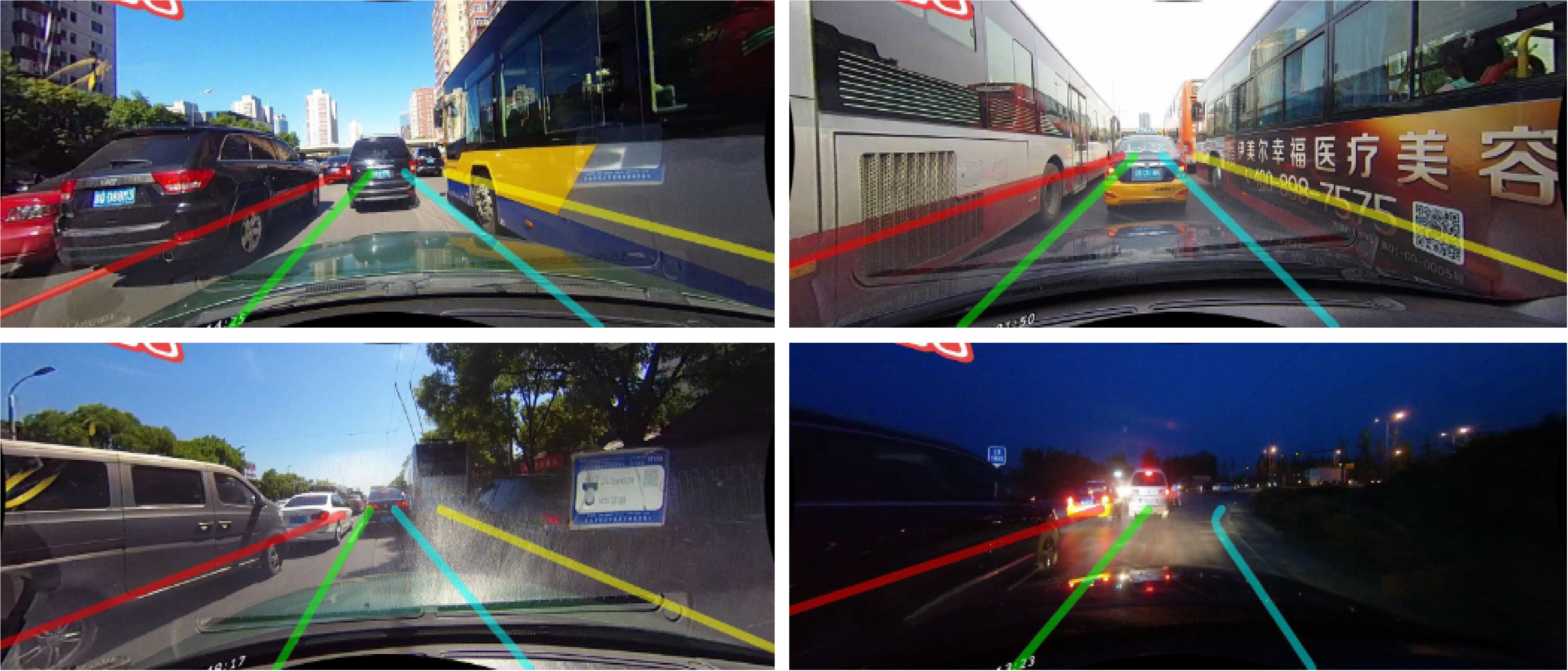}
    \caption{Illustration of difficulties in lane detection. Different lanes are marked with different colors. Most challenging scenarios are severely occluded or distorted with various lighting conditions, resulting in little or no visual clues of lanes that can be used for lane detection.}
    \label{fig_difficult}
    \end{figure}

Besides the efficiency problem, another challenge is the no-visual-clue problem, as shown in \Cref{fig_difficult}. The lane detection task \cite{SCNN,xu2020curvelane} is to find the location of lanes, no matter whether the lanes are visible or not. In this way, how to handle the scenarios with severe occlusions and extreme lighting conditions that have no visible information is a major difficulty in the lane detection task. To alleviate this problem, additional clues that could potentially hint at the detection are crucial. For example, road shape, cars heading direction trends, non-occluded endpoints of lanes, etc., could benefit the detection. In order to make it possible to use the additional clues, enlarging the receptive field to utilize more information is preferable to lane detection. 

\begin{figure*}[t]
    \centering
    \includegraphics[width=0.95\linewidth]{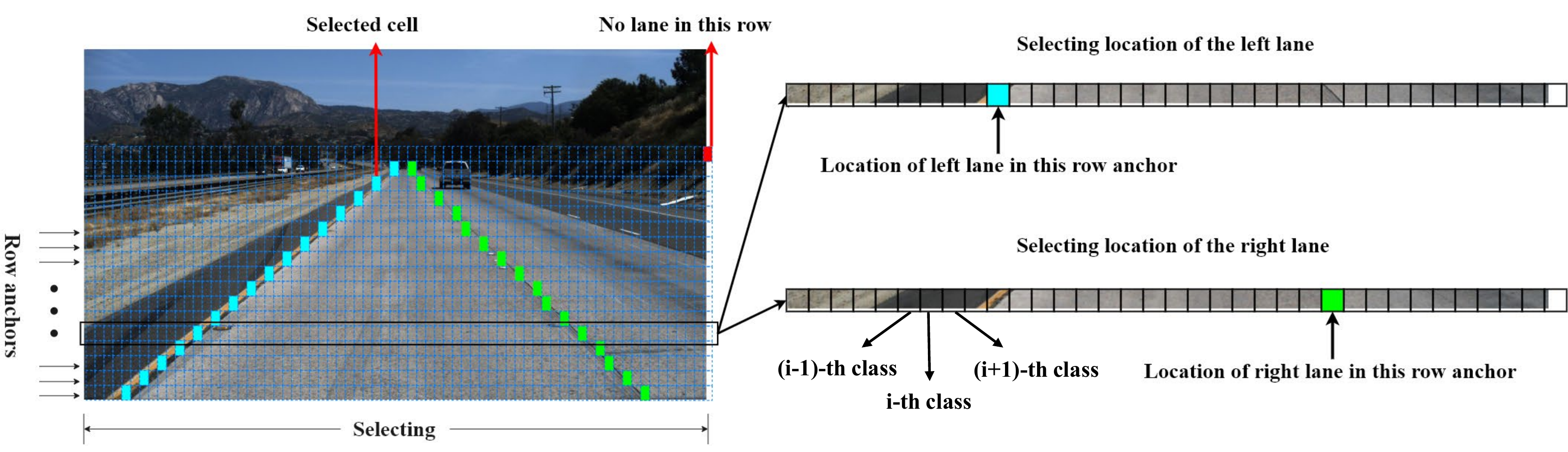}
    \caption{Illustration of the row anchor system. Row anchors are the predefined sparse row locations, and lanes can be represented as coordinates (in a classification-based manner) on the row anchors. }
    \label{fig_row_anchor}
    \end{figure*}

This raises a natural question: can we find a fast and global formulation having large receptive field for the lane detection task?
With the above motivations, we propose a sparse top-down formulation to address the efficiency and no-visual-clue problems. First, we propose a novel row-anchor-driven representation for lanes. A lane can be represented by the coordinates on a series of predefined row anchors, as shown in \Cref{fig_row_anchor}. Since a lane can be well represented by a small set of key points (in a fixed sparse row anchor system), the efficiency problem can be addressed with the sparsity of the anchor-driven representation. Second, we propose to use a classification-based manner to learn the coordinates of lanes with the anchor-driven representation. Using the classification-based manner (working with the whole global feature), the receptive field is as large as the whole input. It enables the network to better capture the global and long-range information for lane detection, and the problem of no-visual-clue can be relieved effectively. 


\begin{figure*}[!t]
	\centering
	\subfloat[Lane definition]{\includegraphics[width=2in]{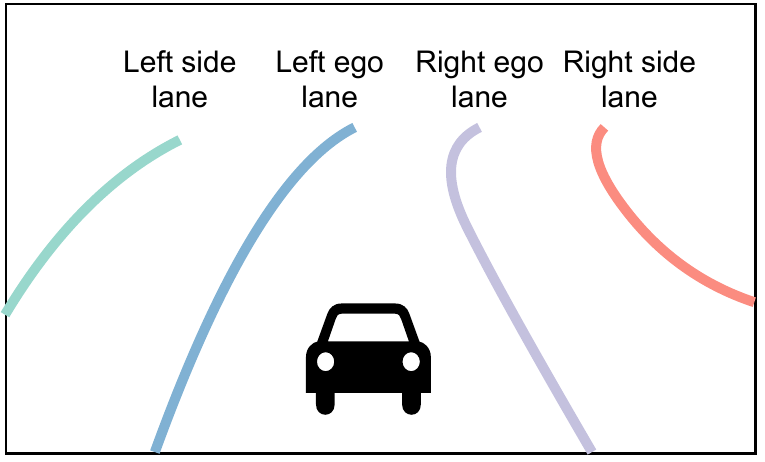}%
    \label{fig_class_rep}}
    \hfil
	\subfloat[Lane-wise Acc. with row anchor]{\includegraphics[width=2in]{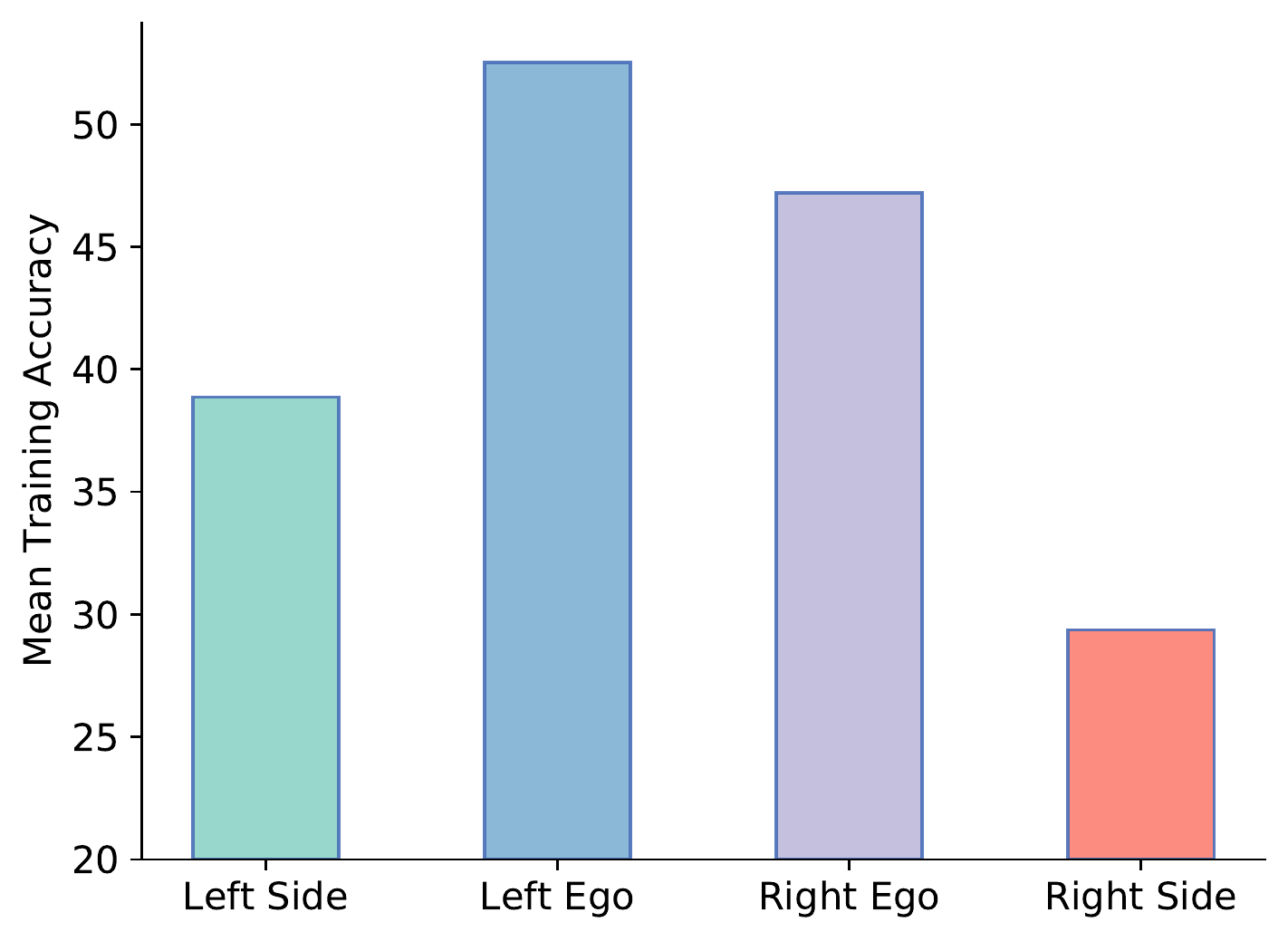}%
    \label{fig_lane_wise_acc_row}}
    \hfil
    \subfloat[Lane-wise Acc. with column anchor]{\includegraphics[width=2in]{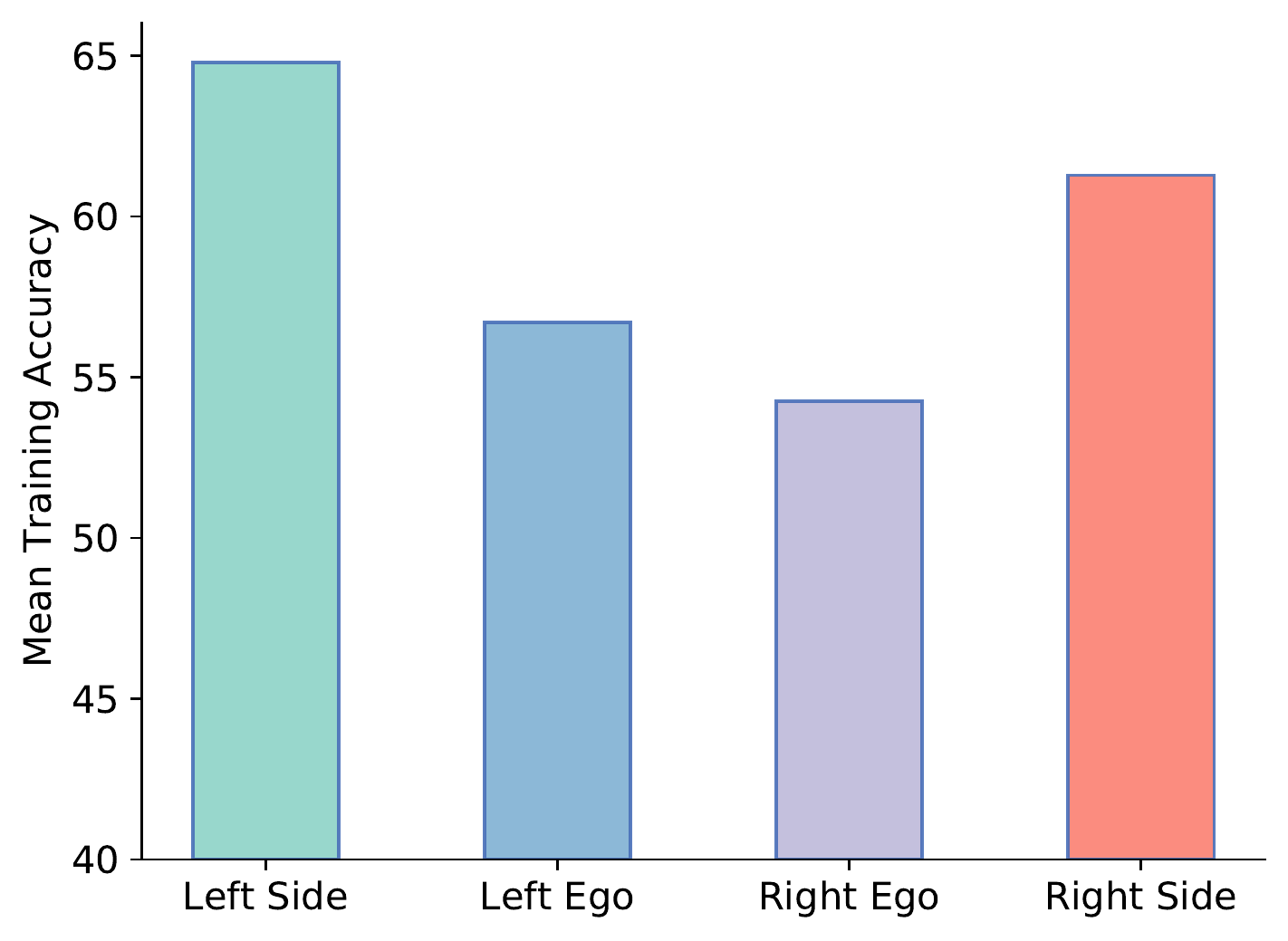}%
    \label{fig_lane_wise_acc_col}}
	\caption{Illustration of the localization accuracy for each kind of lane. (a) shows the definition of lanes by the CULane dataset \cite{SCNN}. (b) is the lane-wise accuracy with the row anchor system. (c) illustrates the lane-wise accuracy with the column anchor system. We can see that for ego lanes, the row anchor system gains better performance, while the column anchor system gains better performance for side lanes.}
	\label{fig_magnified}
\end{figure*}

Furthermore, we extend the lane representation with row anchor to a hybrid anchor system in this work.
Our observation is that the row anchor systems may not work well for all kinds of lanes, and it could cause a magnified localization problem. As shown in \Cref{fig_class_rep,fig_lane_wise_acc_row}, the localization accuracies for side lanes are significantly lower than those for the ego lanes when using row anchors. What if we use column anchors? In \Cref{fig_lane_wise_acc_col}, we can see an opposite phenomenon that the column anchor system has worse localization ability for ego lanes. We name this phenomenon \textit{magnified localization error problem}. This problem makes row anchors difficult to localize the horizontal (side lanes) lanes and similarly makes column anchors hard to localize the vertical lanes (ego lanes). With the above observation, we propose to use hybrid (row and column) anchors to represent different lanes separately. Specifically, we use row anchor for ego lanes and column anchor for side lanes. In this way, the magnified localization error problem can be relieved, and the performance can be improved.

With the hybrid anchor system, a lane can be represented by the coordinates on the anchor system. How to effectively learn these coordinates is another important problem. The most straightforward method is to use regression. Usually, the regression methods work for a local range prediction setting \cite{7410526, 7485869, redmon2016you, liu2016ssd, tian2021fcos, 6619290, zhang2014facial, 7553523} and are relatively weak in modeling long-range and global localization \cite{8765346, wei2016convolutional, newell2016stacked}. To cope with global range prediction, we propose to learn the coordinates of lanes in a classification-based manner, which represents different coordinates using different classes. 
In this work, we further extend the original classification to the ordinal classification. In the ordinal classification, adjacent classes have close and ordinal relationships, which is different from the original classification. For example, in the ImageNet \cite{deng2009imagenet} classification task, the 7th class is stingray (a kind of fish) while the 8th class is cock. In our work, classes are ordered (e.g., the lane coordinates for the 8th class are always spatially on the right of those for the 7th class). Another property of ordinal classification is that the space of classes are continuous. For example, non-integer class like 7.5th class is meaningful, and it can be regarded as an intermediate class between 7th and 8th classes.
To realize ordinal classification, we propose two loss functions together to modeling the ordinal relationship between classes, including a base classification loss and a mathematical expectation loss. Utilizing the ordinal relationship and continuous class space properties, we can use the mathematical expectation instead of argmax to get the continuous class \cite{7406390} of prediction. The expectation loss is to constrain the predicted continuous class to equal the ground truth. 
By simultaneously constraining the base and expectation losses, the output could have a better ordinal relationship and benefit the localization of lanes.

In summary, the main contributions of our work are threefold.
1) We propose a novel, simple, yet effective formulation for lane detection. Compared with the previous methods, our method represents lanes as anchor-based coordinates, and the coordinates are learned in a classification-based manner. This formulation is ultra-fast and effective in solving the no-visual-clue problem. 2) Based on the proposed formulation, we propose a hybrid anchor system, which further extends the previous row anchor system and could effectively reduce the localization error. Moreover, classification-based learning is further extended to an ordinal classification problem, which utilizes the natural ordinal relationship in classification-based localization. 3) The proposed method achieves state-of-the-art speed and performance. Our fastest model could achieve 300+ FPS with comparable performance to the state-of-the-art.


This paper is an extension of our previous conference publication \cite{qin2020ultra}. Compared with the conference version, this paper has extensions as follows:
\begin{itemize}
    \item \textit{Hybrid Anchor System} With the observation of the magnified error problem, we propose a new hybrid anchor system that could effectively reduce localization error compared with the previous publication.
    \item \textit{Ordinal Classification Losses} We propose the new loss functions that treat lane localization as an ordinal classification problem, which further improves the performance.
    \item \textit{Presentation \& Experiments} Most of the paper is rewritten to give clearer presentations and illustrations. We provide more analyses, visualizations, and results to better cover the space of our work. Stronger results with 6.3 points of performance improvement at the same speed are also provided in this version.
\end{itemize}

\section{Related Work}
\label{sec_related}

\subsection{Lane Detection with Bottom-up Modeling}

Traditional approaches usually tackle the lane detection problem using low-level image processing techniques. By using low-level image processing, traditional methods work in a bottom-up manner inherently. Their main idea is to take advantage of visual clues through image processing like the HSI color model \cite{sun2006hsi} and edge extraction algorithms \cite{yu1997lane,wang2000lane}. Gold \cite{bertozzi1998gold} is one of the earliest attempt that detects lanes and obstacles using edge extraction algorithms from a stereo vision system. Besides using the features from different color models and edge extraction methods, \cite{aly2008real} propose to use projective geometry and inverse perspective mapping to utilize the prior information that lanes are typically parallel in the real world. Although many methods try different traditional features for lanes, the semantic information from the low-level image processing is still relatively insufficient in the complex scenarios. In this way, tracking is another popular post-processing solution \cite{wang2004lane,kim2008robust} to enhance the robustness. Besides tracking, markov and conditional random fields \cite{CRF} are also used as post-processing methods. Also, some methods \cite{kluge1995deformable,gonzalez2000lane,mandalia2005using} adopting learning mechanisms (like template matching, decision tree, and support vector machine) are proposed. 

With the development of deep learning, some methods \cite{kim2014robust,Empirical_Evaluation} based on deep neural networks show the superiority in lane detection. These methods usually tackle the lane detection task using heat maps that indicate the existence and locations of lanes. Following these early attempts, the mainstream methods start to treat lane detection as a segmentation problem. For instance, VPGNet \cite{Lee_2017_ICCV} proposes a multi-task segmentation network guided by vanishing points for lanes and road markings detection. To enlarge the receptive field of pixel-wise segmentation and boost the performance, SCNN \cite{SCNN} utilizes a special convolution operation in the segmentation module. It aggregates information from different dimensions via processing sliced features and adding them together one by one, which is similar to recurrent neural networks. RONELD \cite{chng2020roneld} proposes an enhancement method for SCNN by finding and constructing straight and curved active lanes separately. RESA \cite{zheng2020resa} also proposes a similar approach by recurrent feature-shift to enlarge the receptive field.  Due to the heavy computational load of segmentation methods, some works try to explore lightweight methods for real-time applications. Self-attention distillation (SAD) \cite{SAD} applies an attention distillation mechanism, in which high and low layers' attentions are respectively treated as teachers and students. IntRA-KD \cite{hou2020interregion} also uses inter-region affinity distillation to boost the performance of the student network. In this way, a shallow network could have a similar performance as a deep network by attention distillation. CurveLane-NAS \cite{xu2020curvelane} introduces neural architecture search techniques to search segmentation networks tailored for lane detection. In LaneAF \cite{abualsaud2021laneaf}, this work proposes to detect lanes by voting in affinity fields based on the segmentation. FOLOLane \cite{qu2021focus} proposes to model local patterns and achieves global prediction of global structure with a global geometry decoder in a bottom-up manner. 

\subsection{Lane Detection with Top-down Modeling}
Besides the mainstream segmentation formulation, some works also try to explore other formulations for lane detection. In \cite{li2016deep}, a long short-term memory (LSTM) network is adopted to deal with the long line structure of lanes. With the same principle, Fast-Draw \cite{FastDraw} predicts the direction of lanes at each lane point, then draws them out sequentially. In \cite{Proposal-Free}, the problem of lane detection is regarded as instance segmentation by clustering binary segments. E2E \cite{wvangansbeke_2019} proposes to detect lanes by differentiable least-squares fitting and directly predicts the polynomial coefficients of lanes. Similarly, Polylanenet \cite{tabelini2021polylanenet} and LSTR \cite{LSTR} also propose to predict the lanes' polynomial coefficients by deep polynomial regression and Transformer \cite{NIPS2017_3f5ee243} respectively. LaneATT \cite{tabelini2021cvpr} proposes to use object detection pipelines that regard lines in the image as anchors. Then it classifies and localizes lanes from the dense line anchors. 
Following the idea of utilizing vanishing points prior \cite{Lee_2017_ICCV} and object detection pipelines \cite{tabelini2021cvpr}, \cite{su2021structure} proposes to use line anchors guided by vanishing points.  Different from the 2D view of previous works, there are also many methods \cite{garnett20193d, guo2020gen, efrat2020semi, sela20203d} trying to detect lanes in 3D. 

Different from the previous bottom-up works, our method is a top-down modeling method. By top-down modeling, the method could naturally focus more on the global information, which is beneficial to the no-visual-clue problem. Compared with the previous top-down methods, our method is aiming at a new lane detection formulation with the row and hybrid anchor lane representations, which could greatly reduce the learning difficulty and speed-up detection. With the proposed formulation in the previous conference version \cite{qin2020ultra}, our works have been successfully adopted and generalized to other methods\cite{liu2021condlanenet}.
\section{Ultra Fast Lane Detection}
\label{sec_method}
In this section, we describe the details of our method. First, we demonstrate the representation of lanes in coordinates on the proposed hybrid anchor system. Second, we show the design of deep network architecture and corresponding ordinal classification losses. Last, the complexity analysis is elaborated.

\subsection{Lane Representation with Anchors}
\label{sec_hybrid_anchor}

\begin{figure*}[!t]
    \centering
    \includegraphics[width=0.95\textwidth]{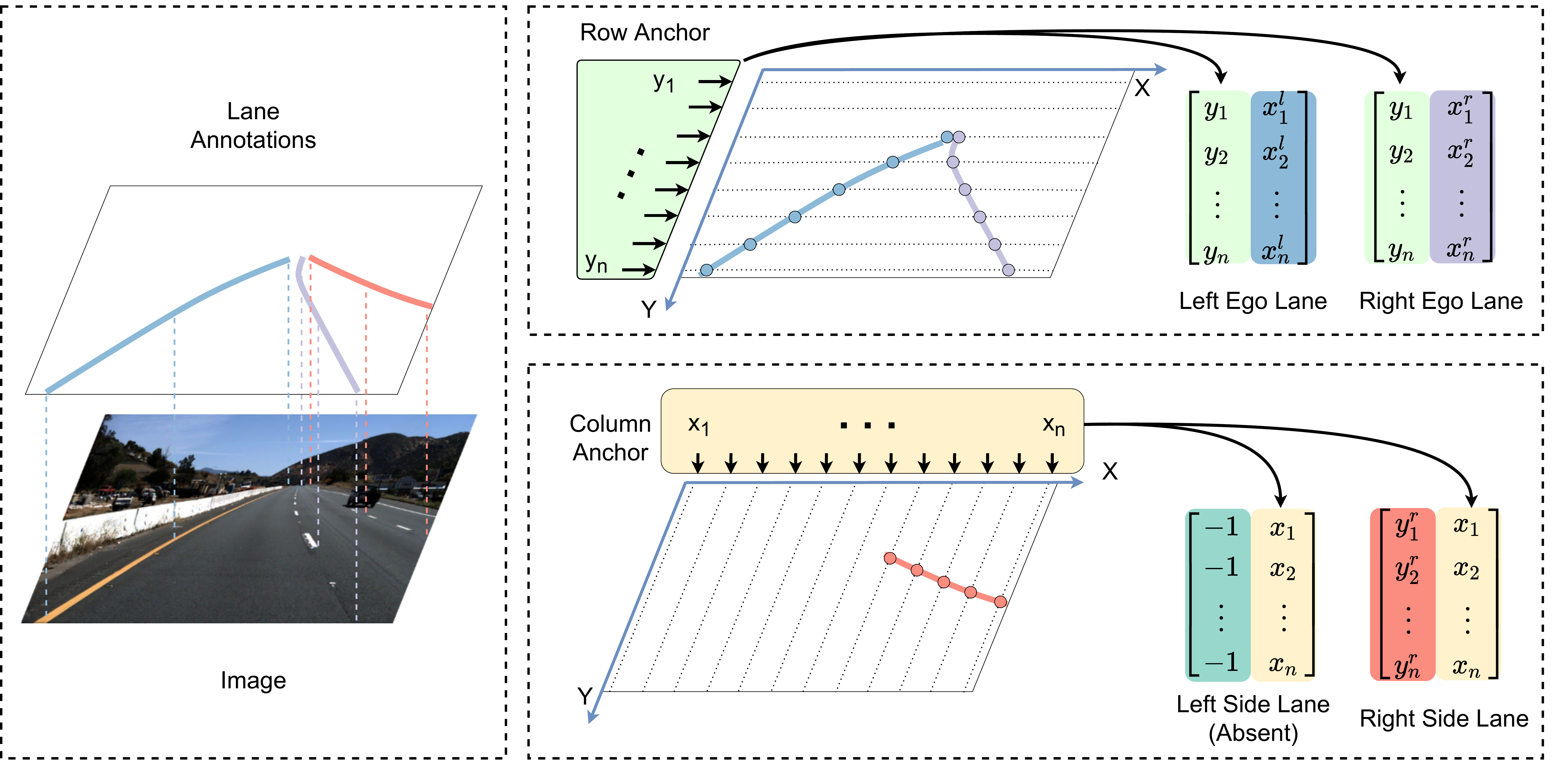}
    \caption{Lane representation with the hybrid anchor system. The input image and its annotations are shown on the left. For the ego lanes, we use row anchors to represent the location of lane markings. For side lanes, the column anchors are adopted. With the hybrid anchor system, the lanes can be simply represented as four matrices corresponding to the four lanes as shown on the right. Moreover, the fixed and shared coordinates of anchors can be left out. For the absent locations, we use -1 to indicate them.}
    \label{fig_form}
\end{figure*}

In order to represent lanes, we introduce the row anchors for lane detection, as shown in \Cref{fig_row_anchor}. Lanes are represented with points on the row anchors.
However, the row anchor system might cause a magnified localization error problem, as shown in \Cref{fig_magnified}. In this way, we further extend the row anchor system to a hybrid anchor system.

\begin{figure}[h]
    \centering
    \includegraphics[width=0.95\linewidth]{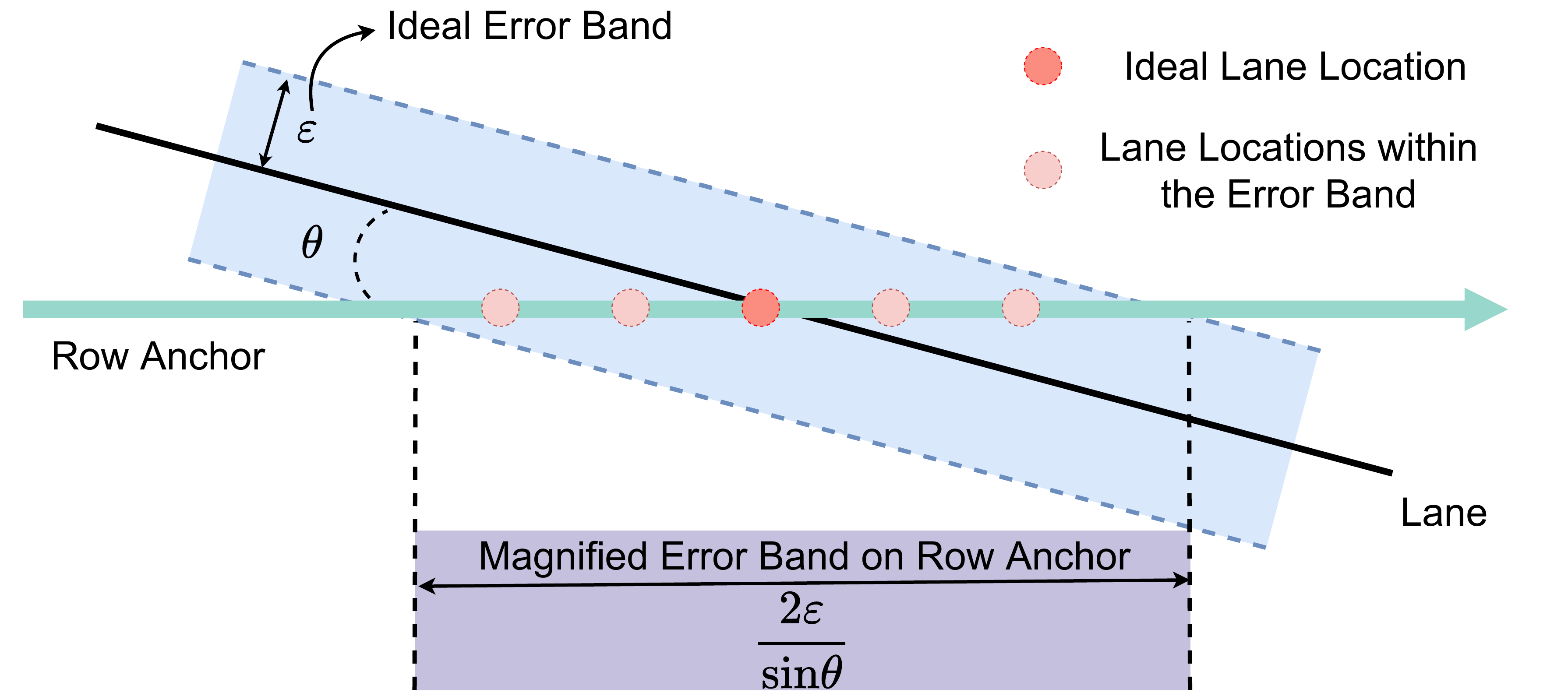}
    \caption{Illustration of the magnified localization error problem. The blue area represents the ideal localization error band (the minimal inevitable error that a network could have. It is possibly caused by the bias of a network, the error introduced by the annotator, etc.) without an anchor system. The purple area shows the magnified error band introduced by the anchor system. We can see that when the angles between lanes and anchors are small, the localization error can be magnified tremendously. The magnified localization error problem also holds for column anchors.}
    \label{fig_error}
    \end{figure}

The reason for this problem is shown in \Cref{fig_error}. Suppose the ideal minimal localization error without any anchor system is $\varepsilon$, which can be caused by network bias, annotation error, etc. We can see that the error band on the row anchor system is multiplied by a factor of $\frac{1}{\sin\theta} $. When the angle $\theta$ between the lane and anchor is small, the magnifying factor $\frac{1}{\sin\theta}$ would go to infinity. For example, when a lane is strictly horizontal, it is impossible to represent the lane with the row anchor system. This problem makes the row anchors hard to localize the lanes which are more horizontal (typically side lanes), and similarly, it makes the column anchors hard to localize the lanes which are more vertical (typically ego lanes). Instead, when the lanes and anchors are vertical, the error introduced by the anchor system is minimal ($\theta=0$ in this case), and it equals the ideal localization error $\varepsilon$. 

Motivated by the above observations, we further propose to use hybrid anchors to represent lanes. For different kinds of lanes, we use different anchor systems to reduce the magnified localization error. Specifically, the rule is that: a lane can only be assigned with one kind of anchor, and the more vertical anchor type to the lane is chosen. In practice, lane detection datasets like CULane \cite{SCNN} and TuSimple \cite{tusimple} only annotate the two ego lanes and the two side lanes, as shown in \Cref{fig_class_rep}. In this way, we use row anchors for ego lanes and column anchors for side lanes, and the magnified localization error problem can be relieved by the hybrid anchor system.

With the hybrid anchor system, we can represent lanes as a series of coordinates on the anchor, as shown in \Cref{fig_form}. Denote $N_{row}$ as the number of row anchors and $N_{col}$ as the number of column anchors. For each lane, we first assign the corresponding anchor system, which has the minimal localization error. Then we calculate the line–line intersection between the lane and each anchor, and record the coordinate of the intersection point. If the lane has no intersection between certain anchors, the coordinates will be set to -1. Suppose the number of lanes assigned to row anchors is $N^r_{lane}$ and the one for column anchors is $N^c_{lane}$. \red{The lanes in an image can be represented by a fixed-size target $T$ where every element is either the coordinate of lane or -1, and its length is $N_{row} \times N^r_{lane} + N_{col} \times N^c_{lane}$. }$T$ can be divided into two parts $T^r$ and $T^c$, which correspond to the parts on row and column anchors, and the sizes are $N_{row} \times N^r_{lane}$ and $N_{col} \times N^c_{lane}$ respectively.


\begin{figure*}[!t]
    \centering
    \includegraphics[width=0.95\textwidth]{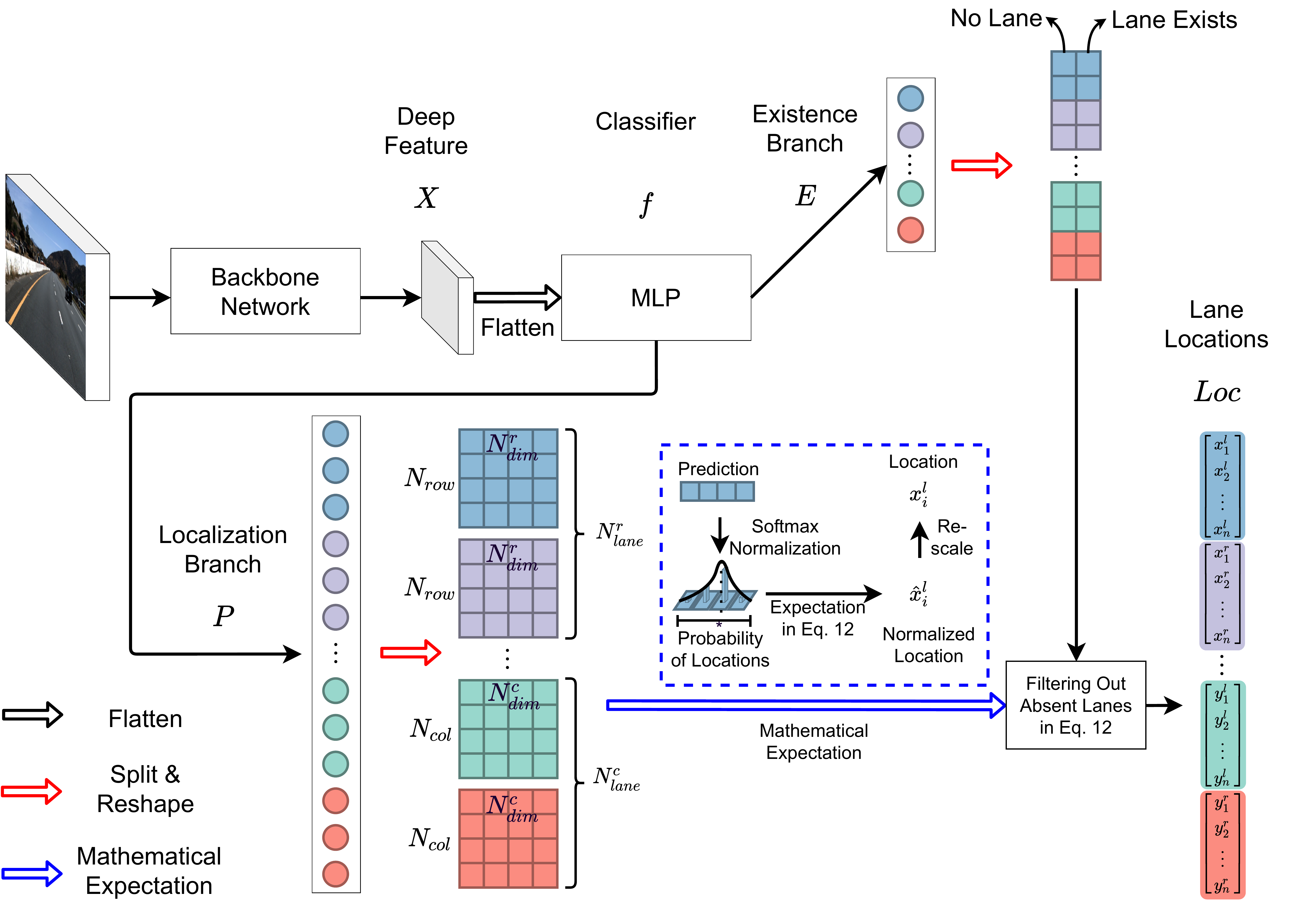}
    \caption{Illustration of the network architecture. The input image is first sent to a backbone network to get the deep feature. Then the deep feature is flattened and fed into a classifier, which has two output branches. The first localization branch is to learn the coordinates on the hybrid anchors with classification-based representation. The second existence branch is to predict the existence of each coordinate on the hybrid anchors. After obtaining the localization output, we use expectation instead of argmax to get the coordinates of lanes.}
    \label{fig_arch}
\end{figure*}

\subsection{Anchor-driven Network Design}

\begin{table}[h]

	\caption{Notations}
	\vspace{-1em}
	\label{tb_notation}
	\setlength{\tabcolsep}{1.8mm}{
		\begin{tabular}{lll}
			\toprule 
			Variable  & Type   & Definition                               \\ \midrule
			$H$         & Scalar & Height of image                          \\
			$W$         & Scalar & Width of image                           \\
			$N_{row}$         & Scalar & Number of row anchors         \\
			$N_{col}$         & Scalar & Number of col anchors         \\
			\specialrule{0em}{2pt}{2pt}
			$N^r_{dim}$         & Scalar & Classification dimension on row anchor \\
			\specialrule{0em}{2pt}{2pt}
			$N^c_{dim}$         & Scalar & Classification dimension on col anchor\\
			\specialrule{0em}{2pt}{2pt}
			$N^r_{lane}$         & Scalar & Number of lanes on row anchor                         \\
			\specialrule{0em}{2pt}{2pt}
			$N^c_{lane}$         & Scalar & Number of lanes on col anchor                      \\
			\specialrule{0em}{2pt}{2pt}
			$f$          &Function& The classifier \\
			$X$         &Tensor& The global feature of image \\
			$P$ & Tensor & Prediction of localization branch \\ 
			$E$ & Tensor & Prediction of existence branch\\ 
			$T$ & Tensor & Lane representation with hybrid anchor \\ 
			$T_{cls}$ & Tensor & Learning target of localization branch\\ 
			$T_{ext}$ & Tensor & Learning target of existence branch\\ 
			$Prob$        & Tensor & Probability of each location              \\
			$Loc$     & Tensor & Location of lanes                 \\
			\bottomrule
	\end{tabular}}

	\medskip
	\emph{\footnotesize For variables $P$, $E$, $T$, $T_{cls}$, $T_{ext}$, $Prob$, and $Loc$, a superscript $\cdot^r$, like $P^r$, represents the parts corresponding the row anchor, while $\cdot^c$ corresponds the column anchor.}
\end{table}
With the help of lane representation with hybrid anchor, our goal of designing networks is to learn the fixed-size targets $T^r$ and $T^c$ with classification. To learn $T^r$ and $T^c$ with classification, we map different coordinates in $T^r$ and $T^c$ to distinct classes. Suppose $T^r$ and $T^c$ are normalized (the elements of $T^r$ and $T^c$ range from 0 to 1 or equal -1, i.e., the ``no lane" case), and the numbers of classes are $N_{dim}^r$ and $N_{dim}^c$. The mapping can be written as:
\begin{equation}
    \begin{aligned}
        & \left\{\begin{matrix}
        T_{cls\_i,j}^r = \left \lfloor T_{i,j}^r N_{dim}^r \right \rfloor,
        \\[3mm]
        T_{cls\_m,n}^c = \left \lfloor T_{m,n}^c N_{dim}^c \right \rfloor,
        \end{matrix}\right. \\
        s.t. \quad & i \in \{1, \cdots, N_{row}\}, j \in \{1, \cdots, N_{lane}^r\},\\
        & m \in \{1, \cdots, N_{col}\}, n \in \{1, \cdots, N_{lane}^c\},
    \end{aligned}
    \label{eq_map_to_cls}
\end{equation}
in which $T_{cls}^r$ and $T_{cls}^c$ are the mapped class labels of the coordinates, $\left \lfloor \cdot \right \rfloor$ is the floor operation, and $T_{cls\_i,j}^r$ is the element in the $i$-th row, $j$-th column of $T_{cls}^r$.
In this way, we could convert the learning of coordinate on the hybrid anchor to two classification problems with dimensions of $N_{dim}^r$ and $N_{dim}^c$, respectively. For the no lane case, i.e., $T_{i,j}^r$ or $T_{m,n}^c$ equals -1, we use an additional two-way classification to indicate:
\begin{equation}
    \begin{aligned}
    & T_{ext\_i,j}^r = 
    \begin{cases}
        1, & \text{if   }\   T_{i,j}^r \ne -1 \\
        0, & \text{otherwise}
    \end{cases}, \\
    & s.t. \quad i \in \{1, \cdots, N_{row}\}, j \in \{1, \cdots, N_{lane}^r\},
    \end{aligned}
\end{equation}
in which $T_{ext}^r$ is the class label of the coordinates' existence, and $T_{ext\_i,j}^r$ is the element in the $i$-th row, $j$-th column of $T_{ext}^r$. The existence targets for column anchor $T_{ext}^c$ is similar:
\begin{equation}
    \begin{aligned}
    & T_{ext\_m,n}^c = 
    \begin{cases}
        1, & \text{if   }\   T_{m,n}^c \ne -1 \\
        0, & \text{otherwise}
    \end{cases}, \\
    & s.t. \quad m \in \{1, \cdots, N_{col}\}, n \in \{1, \cdots, N_{lane}^c\}.
    \end{aligned}
\end{equation}

With the above derivation, the whole network is to learn the $T_{cls}^r$, $T_{cls}^c$, $T_{ext}^r$ and $T_{ext}^c$ with two branches, which are localization and existence branches. Suppose the deep feature of an input image is $X$, the network can be written as:
\begin{equation}
    P,E=f(\text{flatten}(X)),
    \label{eq_net_run}
\end{equation}
in which $P$ and $E$ are the localization and existence branches, $f$ is the classifier, and $\text{flatten}(\cdot)$ is the flatten operation. The outputs of $P$ and $E$ are all composed of two parts ($P^r$, $P^c$, $E^r$ and $E^c$), which correspond to the row and column anchors respectively. The sizes of $P^r$ and $P^c$ are $N_{lane}^r \times N_{row} \times N_{dim}^r$ and $N_{lane}^c \times N_{col} \times N_{dim}^c$ respectively, in which $N_{dim}^r$ and $N_{dim}^c$ are the mapped classification dimensions for row and column anchors. The sizes of $E^r$ and $E^c$ are $N_{lane}^r \times N_{row} \times 2$ and $N_{lane}^c \times N_{col} \times 2$ respectively.

In \Cref{eq_net_run}, we directly flatten the deep features from the backbone and feed them to the classifier. In comparison, conventional classification networks \cite{Simonyan15, szegedy2015going, resnet, xie2017aggregated} use global average pooling (GAP). The reason why we use flatten instead of GAP is that we find the spatial information is crucial for the classification-based lane detection network. Using GAP would eliminate the spatial information and result in poor performance.

\subsection{Ordinal Classification Losses}
\label{sec_ordinal_cls}
As we can see in \Cref{eq_map_to_cls}, an essential property is that the classes in the above classification network have an ordinal relationship. In our classification network, adjacent classes are defined to have a close and ordinal relationship, which is different from conventional classification. To better utilize the prior of the ordinal relationship, we propose to use a base classification loss and an expectation loss.

The base classification loss is defined as:
\begin{equation}
    \begin{aligned}
    L_{cls} = & \sum_{i=1}^{N_{lane}^r} \sum_{j=1}^{N_{row}} L_{CE}(P^r_{i,j}, \text{onehot}(T_{cls\_i,j}^r)) \\
    + & \sum_{m=1}^{N_{lane}^c} \sum_{n=1}^{N_{col}} L_{CE}(P^c_{m,n}, \text{onehot}(T_{cls\_m,n}^c)),
    \end{aligned}
\end{equation}
in which $L_{CE}(\cdot)$ is the cross entropy loss, $P^r_{i,j}$ is the prediction of $i$-th lane that is assigned to row anchor and $j$-th row anchor, $T_{cls\_i,j}^r$ is the corresponding classification label to $P^r_{i,j}$, $P^c_{m,n}$ is the prediction of $m$-th lane that is assigned to column anchor and $n$-th column anchor, $T_{cls\_m,n}^c$ is the corresponding classification label to $P^c_{m,n}$, and $\text{onehot}(\cdot)$ is the one-hot encoding function. 

Since the classes are ordinal, the expectation of the prediction can be regarded as the mean voting results. For convenience, we denote the expectation as:
\begin{equation}
    \left\{\begin{matrix}
        Exp^r_{i,j} \ = \sum_{k=1}^{N_{dim}^r}Prob^r_{i,j}[k] \cdot k,
        \\[2mm]
        Exp^c_{m,n} = \sum_{l=1}^{N_{dim}^c}Prob^c_{m,n}[l] \cdot l,
       \end{matrix}\right.
\end{equation}
in which $[\cdot]$ denotes the indexing operator. The $Prob$ is defined as:

\begin{equation}
    \left\{\begin{matrix}
    Prob_{i,j}^r \ \  = \text{softmax}(P_{i,j}^r),
    \\
    \ Prob_{m,n}^c = \text{softmax}(P_{m,n}^c),
\end{matrix}\right.
\end{equation}

In this way, we could constrain the expectation of prediction to get close to the ground truth. Thus, we have the following expectation loss:
\begin{equation}
    \begin{aligned}
        L_{exp} = & \sum_{i=1}^{N_{lane}^r} \sum_{j=1}^{N_{row}} L_{1}(Exp^r_{i,j}, T_{cls\_i,j}^r)\\
        + & \sum_{m=1}^{N_{lane}^c} \sum_{n=1}^{N_{col}} L_{1}(Exp^c_{m,n}, T_{cls\_m,n}^c),
        \end{aligned}
\end{equation}
in which $L_1(\cdot)$ is the smooth L1 loss. 


\begin{figure}[!t]
    \centering
    \includegraphics[width=0.95\linewidth]{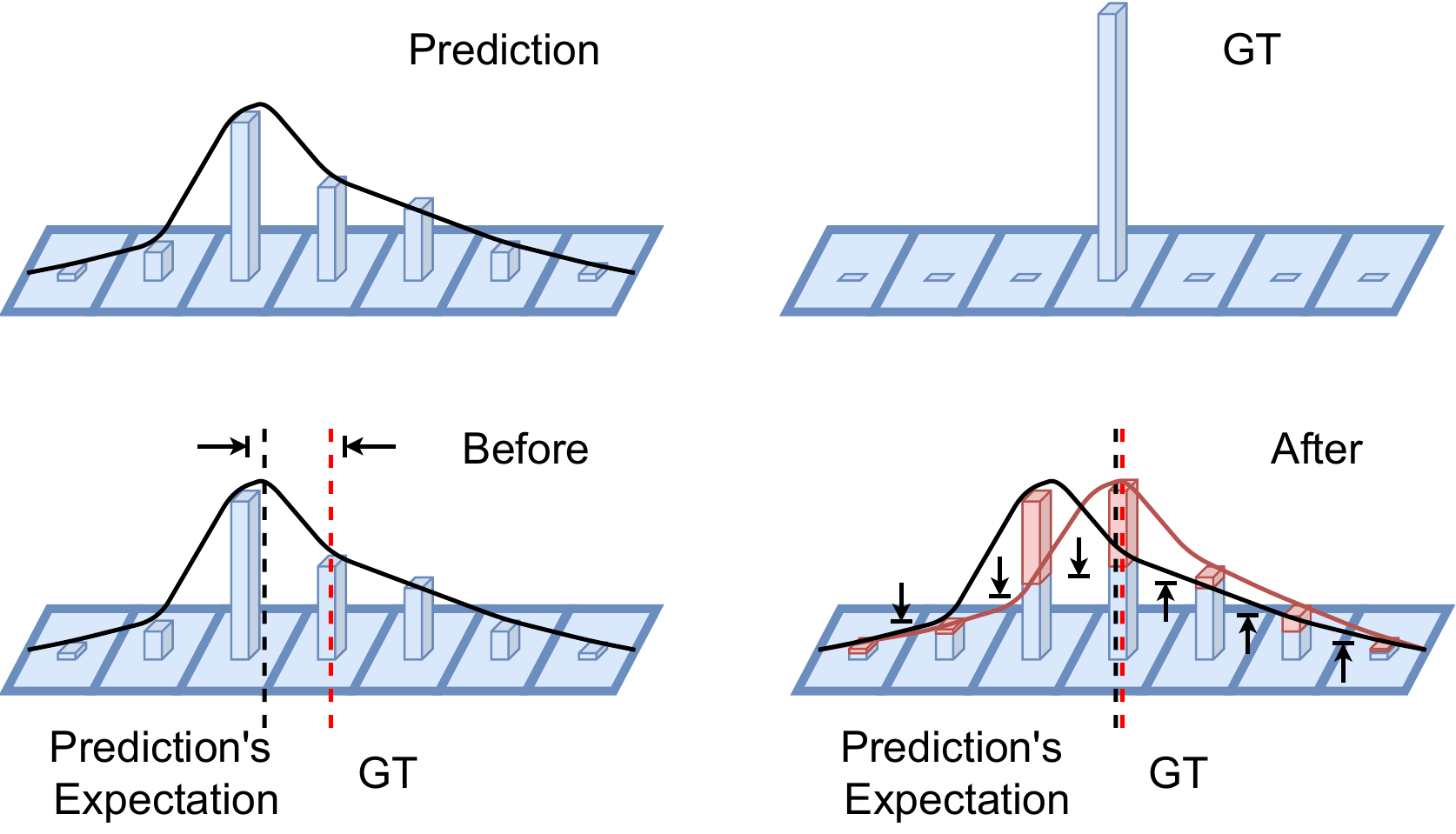}
    \caption{Illustration of the proposed expectation loss.}
    \label{fig_loss}
\end{figure}

The illustration of the expectation loss is shown in \Cref{fig_loss}. We can see that the expectation loss could push the predicted distribution's mathematical expectation towards the location of ground truth, hence benefits the localization of lanes.

In addition, the loss of existence branch $L_{ext}$ is defined as:
\begin{equation}
    \begin{aligned}
        L_{ext} = & \sum_{i=1}^{N_{lane}^r} \sum_{j=1}^{N_{row}} L_{CE}(E^r_{i,j}, \text{onehot}(T_{ext\_i,j}^r))\\
        + & \sum_{m=1}^{N_{lane}^c} \sum_{n=1}^{N_{col}} L_{CE}(E^c_{m,n}, \text{onehot}(T_{ext\_m,n}^c)).
    \end{aligned}
\end{equation}

Finally, the total loss can be written as:
\begin{equation}
    L = L_{cls} + \alpha L_{exp} + \beta L_{ext},
    \label{eq_all}
\end{equation}
in which $\alpha$ and $\beta$ are the loss coefficients.

\subsection{Network Inference}
In this section, we show how to get the detection results during inference. Taking row anchor system as an example, suppose $P_{i,j}^r$ and $E_{i,j}^r$ are the predictions of the $i$-th lane and $j$-th anchor. Then the length of $P_{i,j}^r$ and $E_{i,j}^r$ are $N_{dim}^r$ and 2 respectively. The probability of each location of lanes can be written as:
\begin{equation}
    Prob_{i,j}^r = \text{softmax}(P_{i,j}^r),
\end{equation}
in which the length of $Prob_{i,j}^r$ is $N_{dim}^r$.
Then the location of lanes is obtained with the mathematical expectation of the predicted distribution. Moreover, predictions of absent lanes will be filtered out according to the prediction of the existence branch:
\begin{equation}
    Loc_{i,j}^r = 
    \begin{dcases}
        \sum_{k=1}^{N_{dim}^r} Prob_{i,j}^r[k] \cdot k, & \text{if}\quad E_{i,j}^r[2] > E_{i,j}^r[1] \\
        -1, & \text{otherwise}
    \end{dcases},
    \label{eq_exp}
\end{equation}
Finally, the obtained location $Loc$ is scaled to fit the size of the input image. The overall illustration of the network architecture is shown in \Cref{fig_arch}.

\subsection{Analysis and Discussion}
In this section, we first analyze the complexity of our method and give the reason why our method could achieve ultra-fast speed.

\begin{figure*}[!t]
	\centering
	\subfloat[Our formulation]{\includegraphics[width=4.5in]{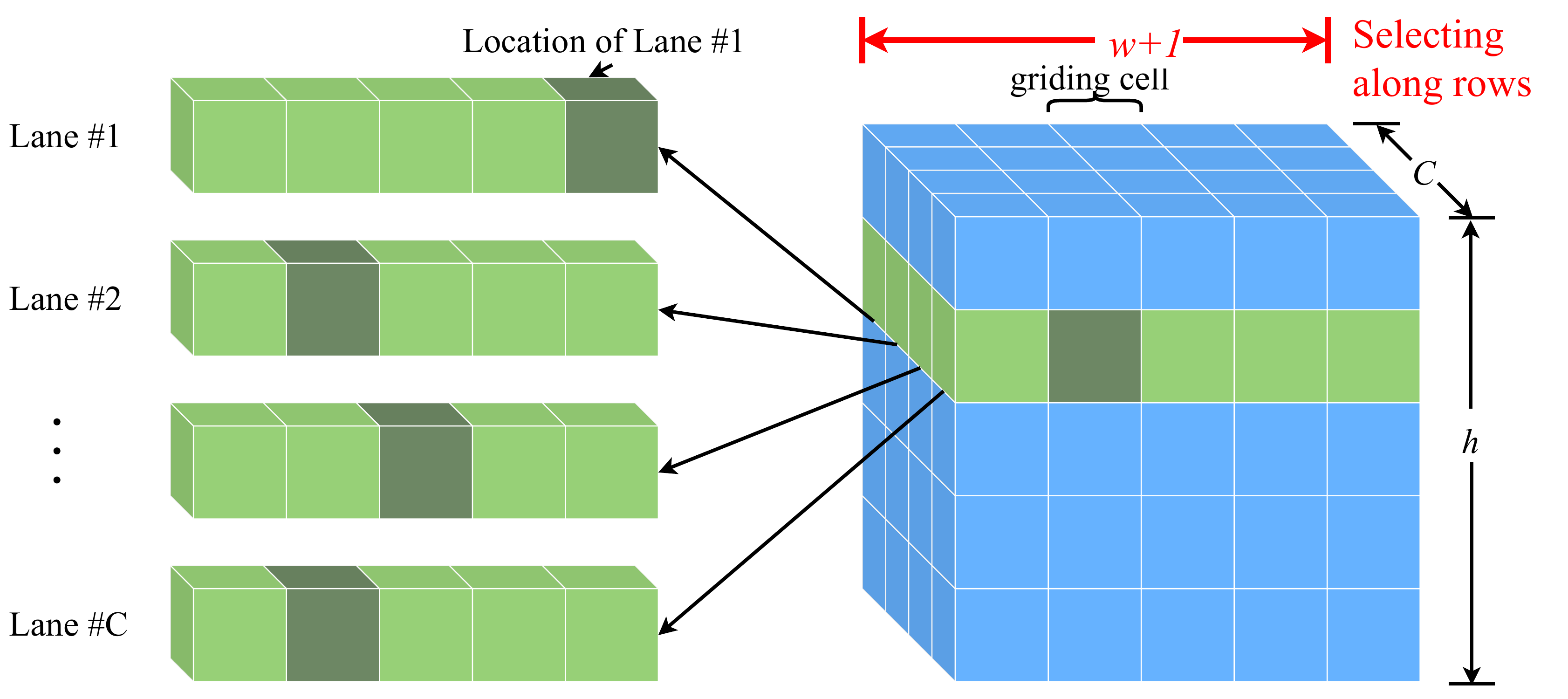}%
    \label{fig_diff_our}}
    \hfil
	\subfloat[Segmentation]{\includegraphics[width=2.325in]{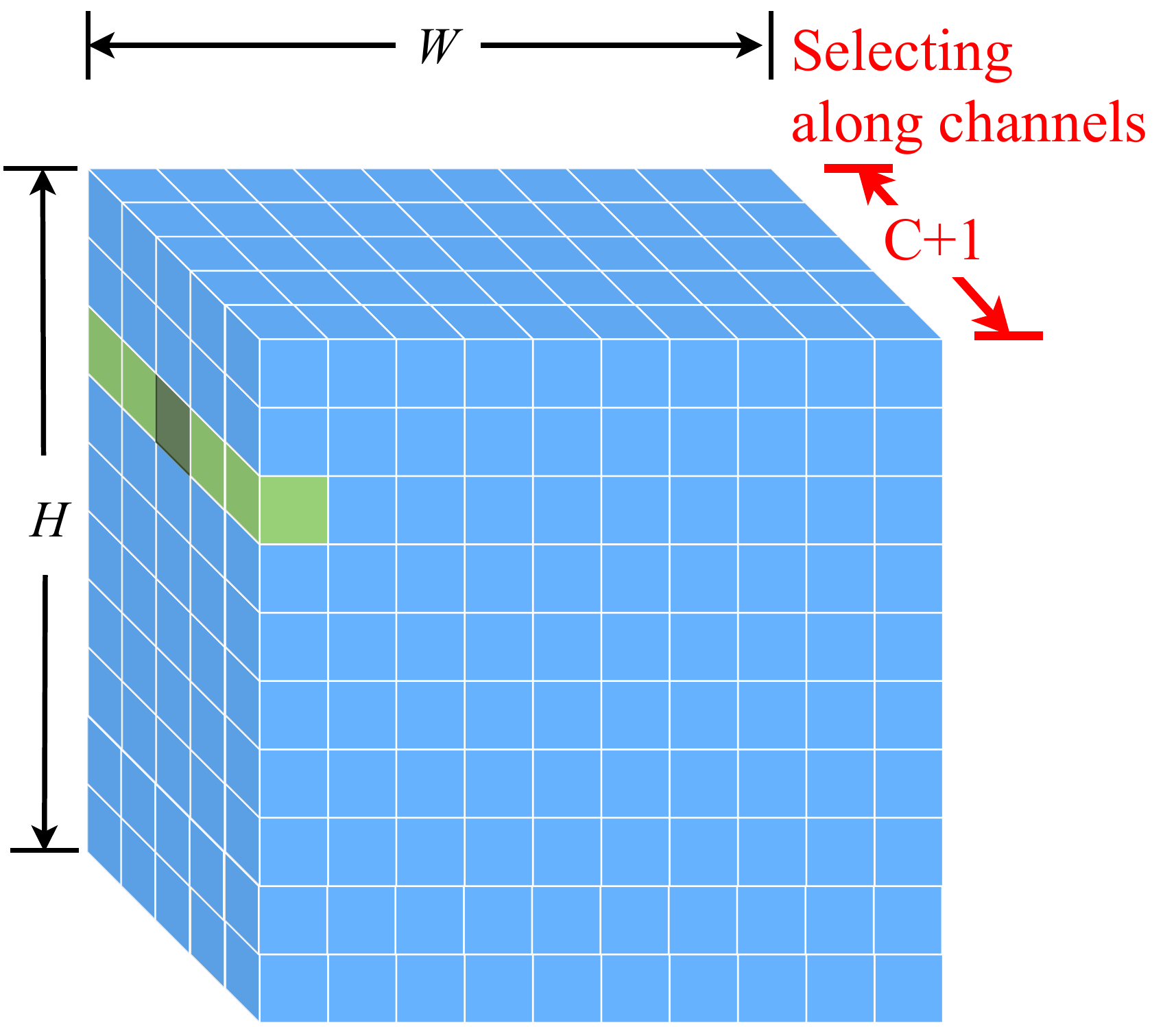}%
    \label{fig_diff_seg}}
	\caption{Illustration of our formulation (row anchor in this case) and conventional segmentation. Taking row anchor as an example; our formulation is selecting locations (grids) on rows, while segmentation is classifying every pixel. The dimensions used for classifying are also different, which is marked with red. The proposed formulation significantly reduces the computational cost. Besides, the proposed formulation uses global feature as input, which has a larger receptive field than segmentation, thus addressing the \textit{no-visual-clue} problem.}
	\label{fig_formulation_diff}
\end{figure*}

In order to analyze the complexity, we use segmentation as a baseline. The differences between our formulation (taking row anchor as an example) and segmentation are shown in \Cref{fig_formulation_diff}. It can be seen that our formulation is much simpler than the commonly used segmentation. Suppose the image size is $H \times W$. Since segmentation is pixel-wise classification, it needs to conduct $H \times W$ classifications. For our method, the length of learning target $T$, which contains the coordinates of lanes on hybrid anchor, is $N_{row} \times N^r_{lane} + N_{col} \times N^c_{lane}$. Since we only need a small number of anchors to represent lanes, and we have $N_{row} \ll H$ and $N_{col} \ll W$. Moreover, $N^r_{lane}$ and $N^c_{lane}$ are the number of lanes assigned to row and column anchors, which is very small compared with other variables. In this way, we have:
\begin{equation}
    \underbrace{N_{row} \times N^r_{lane} + N_{col} \times N^c_{lane}}_{\text{\#CLS of our method}} \ll \underbrace{H \quad \times \quad W}_{\text{\#CLS of the segmentation}},
\end{equation}
in which \#CLS means ``the number of classifications". Taking the settings on the CULane \cite{SCNN} dataset as an example, we have $H=320$, $W=1600$, $N_{row}=18$, $N_{col}=41$, and $N^r_{lane}=N^c_{lane}=2$. Our method's number of classifications is $118$, while the one of segmentation is $5.12\times 10^5$. \red{Taking the classification dimensions into account, the ideal number of calculations of our method is $1.54 \times 10^4$, while the one for segmentation is $2.56\times 10^6$. The computational costs of our classification head and segmentation head with ERFNet~\cite{hou2020interregion,romera2017erfnet} are 0.04 GMac and 30.86 GMac, respectively.}

\begin{figure}[h]
    \centering

    \centering
    \subfloat[Vanilla]{\includegraphics[width=0.45\linewidth]{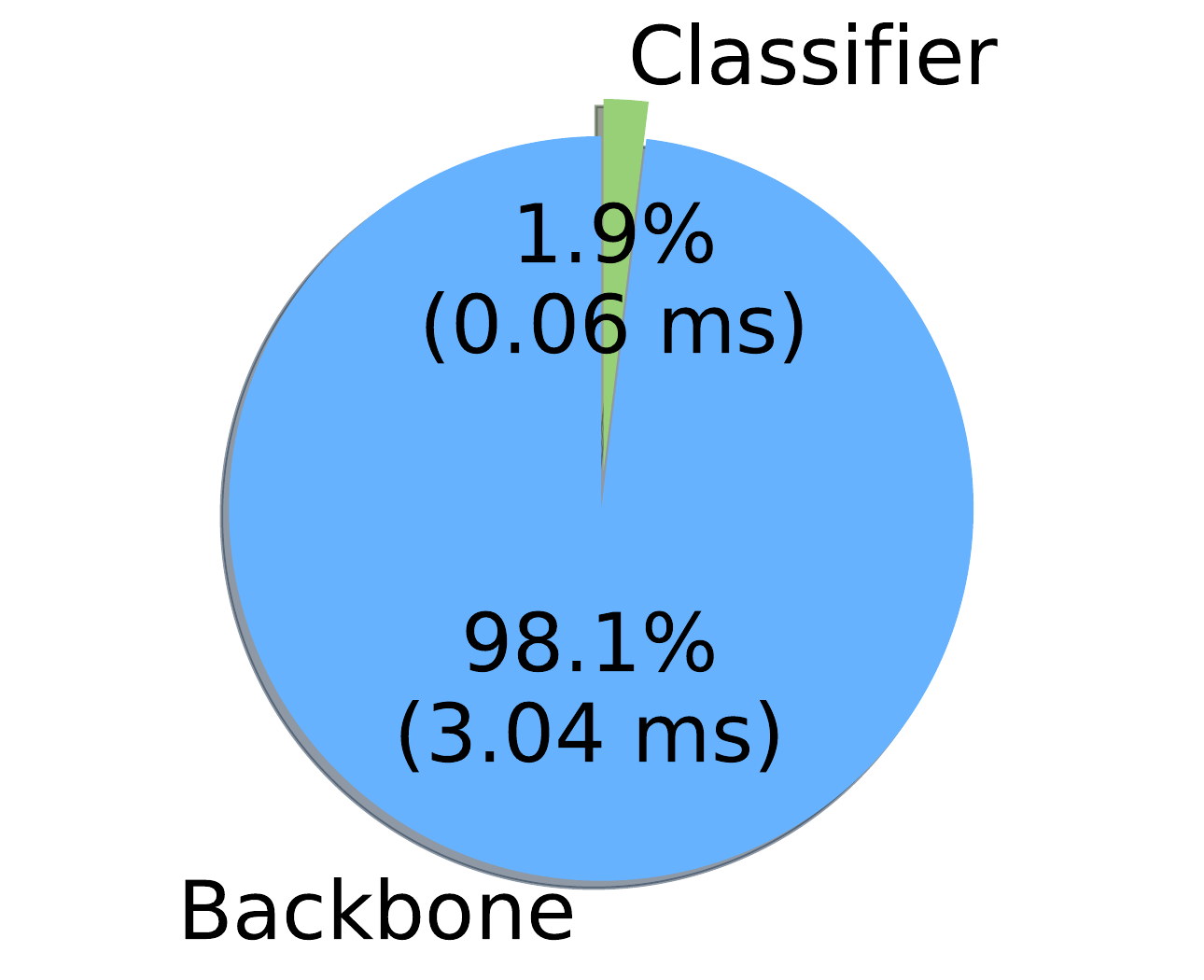}%
    \label{fig_speed_pie_vanilla}}
    \hfil
    \subfloat[W/ FLTTA]{\includegraphics[width=0.45\linewidth]{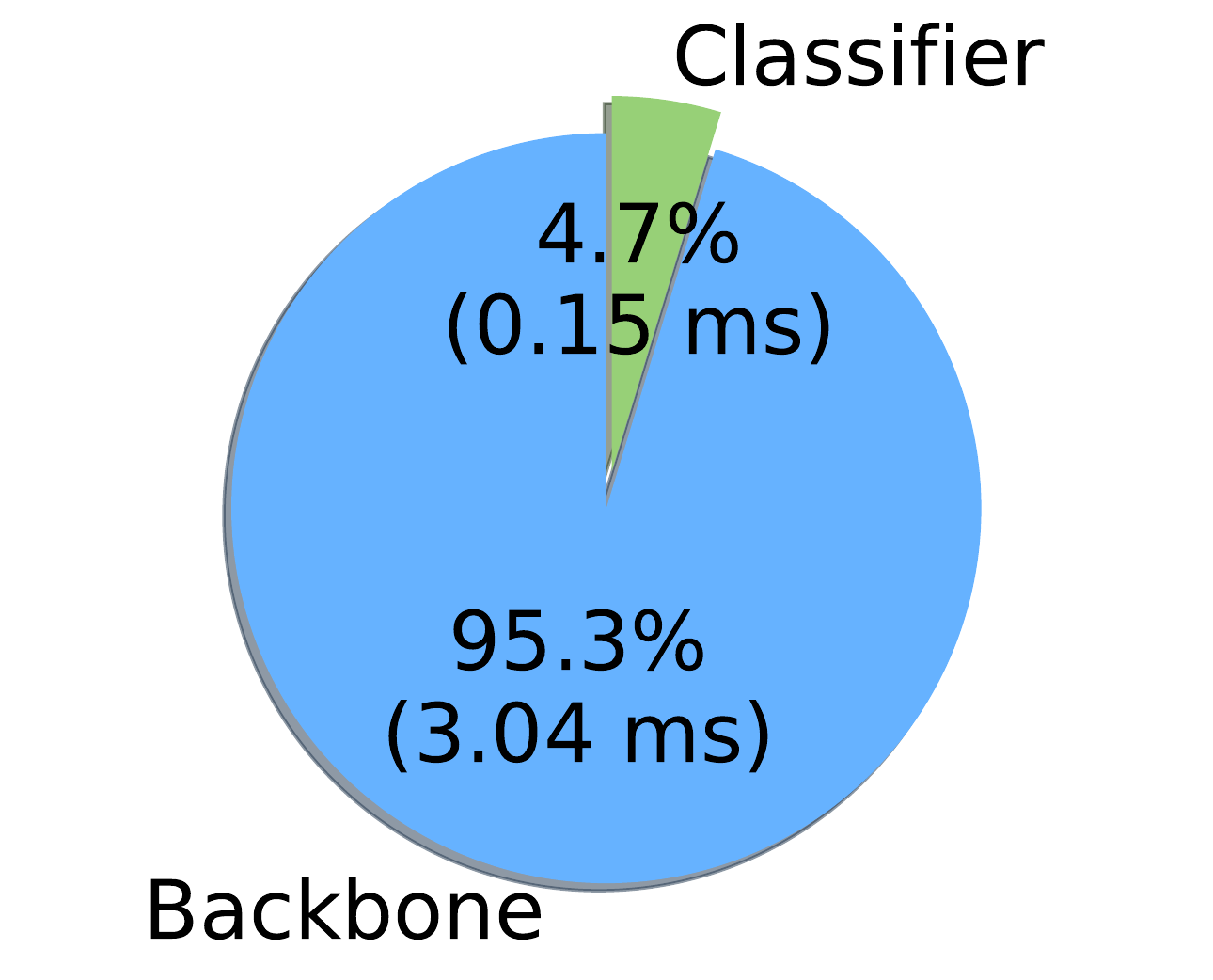}%
    \label{fig_speed_pie_tta}}

    \caption{Pie chart of the forward pass timing for the proposed network with a ResNet-18 backbone. We can see our method is highly efficient, and less than 5\% of computations are from the proposed classifier for lane detection.}
    \label{fig_speed_pie}
\end{figure}

Besides the theoretical analysis, the real forward pass timing of our method is shown in \Cref{fig_speed_pie_vanilla}. We can see that the backbone takes up most of the time. In contrast, the proposed classification-based lane detection head is highly efficient and only costs less than 5\% of the whole inference time.


\section{Experiments \protect\footnote{We provide a visualization demo video for our method. It can be found at: \protect\url{https://youtu.be/VkvpoHlaMe0} and \protect \url{https://www.bilibili.com/video/BV1gV411x7KX/}}}
\label{sec_exp}
In this section, we demonstrate the effectiveness of our method with extensive experiments. The following sections mainly focus on three aspects: 1) Experimental settings. 2) Ablation studies of our method. 3) Results on four major lane detection datasets. 

\begin{table*}[]
	\centering
	\caption{\red{Datasets description}}
	\vspace{-1em}
    \setlength{\tabcolsep}{4mm}{

		\begin{tabular}{ccccccccc}
			\toprule
			Dataset  & \#Frame & Train  & Validation & Test   & Resolution   & \#Lane & \#Scenarios & Environment \\ \midrule
			TuSimple & 6,408   & 3,268  & 358   & 2,782  & 1280$\times$720 & $\leq$5 & 1 & highway     \\
			CULane   & 133,235 & 88,880 & 9,675 & 34,680 & 1640$\times$590 & $\leq$4 & 9 & urban and highway    \\ 
			CurveLanes   & 150,000 & 100,000 & 20,000 & 30,000 & 2560$\times$1440 & $\leq$10 & 1 & urban and highway    \\ 
			LLAMAS   & 100,042 & 58,269 & 20,844 & 20,929 & 1276$\times$717  & $\leq$4 & 1 & highway    \\ 
			
			\bottomrule
	\end{tabular}}
	\label{tab:dataset}

\end{table*}

\subsection{Experimental Setting}
\label{sec_setting}

\subsubsection{Datasets} \red{To evaluate our approach, we conduct experiments on four widely used benchmark datasets: TuSimple \cite{tusimple}, CULane \cite{SCNN}, CurveLanes \cite{xu2020curvelane}, and LLAMAS \cite{llamas2019} datasets. TuSimple dataset is collected with stable lighting conditions on highways. On the contrary, the CULane dataset consists of nine different scenarios, including normal, crowd, curve, dazzle light, night, no line, shadow, and arrow in the urban area. CurveLanes dataset focus on more curved scenarios, and LLAMAS dataset is collected using maps. The detailed information about the datasets can be seen in \Cref{tab:dataset}.}

\subsubsection{Evaluation Metrics} The official evaluation metrics of the datasets are different. For the TuSimple dataset, the evaluation metric is accuracy. The accuracy is calculated by: 
\begin{equation}
accuracy = \dfrac{\sum_{clip}C_{clip}}{\sum_{clip}S_{clip}} ,
\end{equation}
in which $C_{clip}$ is the number of lane points predicted correctly and $S_{clip}$ is the total number of ground truth in each clip. \red{Besides the accuracy, we also use the evaluation metric of CULane for all datasets, in which intersection-over-union (IoU) is computed between ground truth and predictions. }Predictions with IoUs larger than 0.5 are considered as true positives. F1-measure is taken as the evaluation metric and formulated as follows:
\begin{equation}
F1 = \frac{2 \times Precision \times Recall}{Precision + Recall},
\end{equation}
where $Precision = \frac{TP}{TP + FP}$, $Recall = \frac{TP}{TP + FN}$ , $FP$ and $FN$ are the false positive and false negative.

\subsubsection{Implementation Details} The hyperparameter settings are shown in \Cref{tb_hyper}. The corresponding ablation studies of number of anchors and classification dimensions can be seen in \Cref{ab_griding,ab_num_anchor}, respectively.

\begin{table}[]
    \centering
    \caption{\red{Hyperparameter Settings on Different Datasets.}}
    \vspace{-1em}
    \begingroup
    \setlength{\tabcolsep}{2mm}
    \renewcommand{\arraystretch}{1.2}
    \begin{tabular}{ccccc}
    \toprule
      Dataset &Tusimple  & CULane & CurveLanes & LLAMAS \\\midrule
        $N_{row}$ & 56 & 18         & 72 & 36 \\
        $N_{col}$ & 40 & 40         & 80 & 40\\
        $N_{dim}^r$ & 100 & 200     & 200 & 200\\
        $N_{dim}^c$ & 100 & 100     & 100 & 100 \\
        $N_{lane}^r$ & 2  & 2       & 10 & 4 \\
        $N_{lane}^c$ & 2   &2       & 10 & 4\\

    \bottomrule
    \end{tabular}
    \endgroup
    \label{tb_hyper}
\end{table}

In the optimizing process, images are resized to 1600$\times$320 and 800$\times$320 on CULane and TuSimple, respectively. Loss coefficients $\alpha$ and $\beta$ in \Cref{eq_all} are set to 0.05 and 1, respectively. \red{The batch size is set to 16 per GPU, and the total number of training epochs is set to 30 for all datasets. The learning rate is set to 0.1 with SGD optimizer and decreases by a factor of 10 at 25th epoch.} All models are trained and tested with PyTorch \cite{paszke2017automatic} and Nvidia RTX 3090 GPUs.

\subsubsection{Feature Level Test Time Augmentation (FLTTA)}
In this part, we show the test time augmentation method for our method. Because our method works in a global-feature-based classification way, this offers us an opportunity to conduct fast test time augmentation on the feature level. Different from the test time augmentation (TTA) in the object detection, which usually needs to compute the whole backbone repeatedly during TTA, we conduct FLTTA directly on top of the backbone feature. The backbone feature is computed once, and then it is duplicated with spatial shifts in the up, down, left, and right directions. The five augmented feature copies are then fed into the classifier in a batch. Finally, the outputs are integrated to give better predictions during testing. Since the backbone feature is only computed once and the computation of the classifier works in a batched way that takes advantage of the parallel mechanism of GPUs, the FLTTA works nearly as fast as the vanilla testing without TTA. \Cref{fig_speed_pie_tta} shows the forward pass timing pie chart of FLTTA.

\subsubsection{Data Augmentation}
\begin{figure}
	\centering
	\subfloat[Original annotation]{\includegraphics[width=0.45\linewidth]{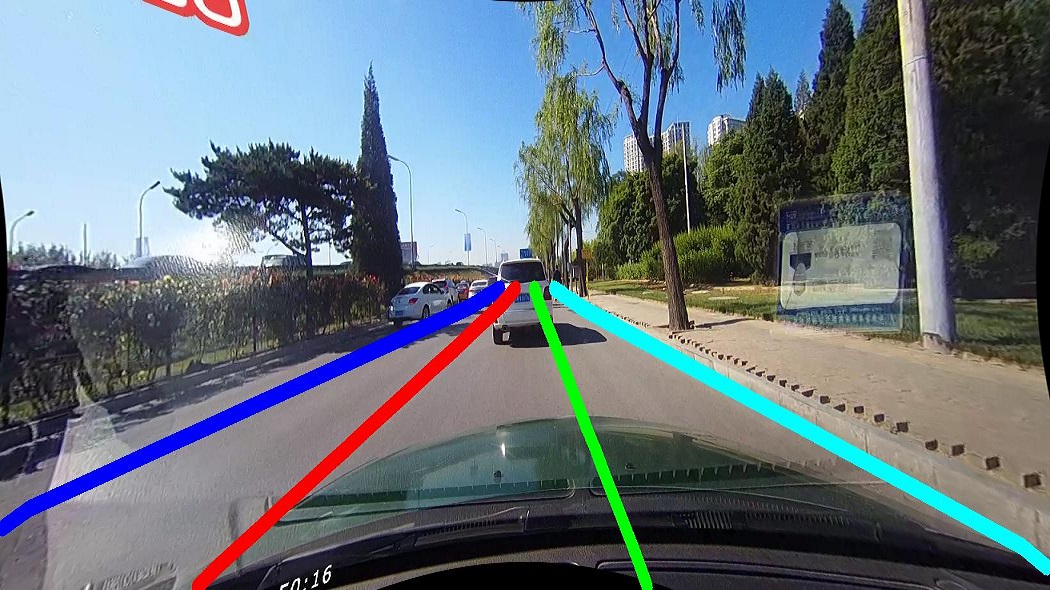}}
    \hfil
	\subfloat[Augmented result]{\includegraphics[width=0.45\linewidth]{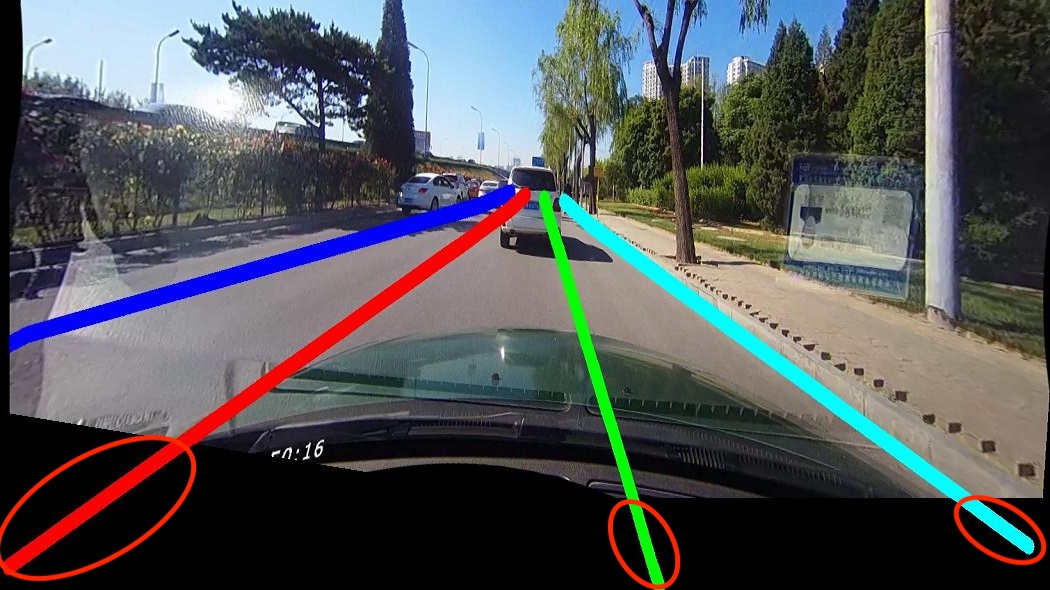}}
	\caption{Demonstration of the spatial shift augmentation. The lane on the right image is extended to maintain the lane structure, which is marked with red ellipse.}
	\label{fig_aug}
\end{figure}
With simple data augmentations like resize and crop, the proposed method could rapidly overfit the whole training set (nearly 100\% training accuracy while poor performance on the testing set). In order to overcome the overfitting problem, we propose a spatial shift data augmentation, which randomly moves the whole image and lane annotations spatially to make the network learn the diverse spatial patterns of lanes. Since some parts of the image and lane annotations are cropped after the spatial shift, we extend the lane annotations until the image boundary. \Cref{fig_aug} shows the augmentation.

\subsection{Ablation Study}
\label{sec_ablation}
In this section, we verify our method with several ablation studies. The experiments are all conducted with the same settings as \Cref{sec_setting}. All experiments for ablation study are conducted with the backbone of the ResNet-18 network.

\subsubsection{Effectiveness of the Hybrid Anchor System}
As we have stated in \Cref{sec_hybrid_anchor}, the row anchor and column anchor system play different roles for lane detection. Motivated by this, we propose a hybrid anchor system, which assigns different lanes to the corresponding anchor type. 

To verify the effectiveness of the hybrid anchor systems, we conduct three experiments on CULane. The results are shown in \Cref{tb_hybrid_anchor}.

\begin{table}[h]

	\caption{Comparison between different anchor systems.}
	\label{tb_hybrid_anchor}

	\vspace{-1em}
	\setlength{\tabcolsep}{7mm}{
	\begin{tabular}{lccc}
	\toprule
	Dataset     & Row & Col & Hybrid \\ \midrule
	CULane      & 66.09 \protect\tablefootnote{\red{The performance of Row is slightly lower than the original version, which is caused by the post-processing. Details can be found in the supplementary materials. }}  & 57.29  & 74.70         \\
	\bottomrule
    \end{tabular}}

	\medskip
 	\emph{\footnotesize Results are based on ResNet-18. Row means only using row anchor, Col means only using column anchor, and Hybrid corresponds to the hybrid anchor. F1-measure is used for evaluation.}
\end{table}

We can see that the hybrid anchor system brings significant improvement compared with row anchors and column anchors, and it shows the effectiveness of hybrid anchors.

\subsubsection{Effectiveness of the Ordinal Classification}
Our method formulates lane detection as an ordinal classification problem. One natural question is that how about other methods like regression and conventional classification.

For the regression method, we replace the classifier head in our pipeline with a similar regression head. The training loss is replaced with Smooth L1 loss. For the conventional classification method, we use the same classifier head as the one in our pipeline. The differences between classification and ordinal classification are the losses and post-processing. \red{1) Classification setting only uses cross-entropy loss while ordinal classification setting uses all three losses as in \Cref{eq_all}.} 2) Classification setting use argmax as its standard post-processing while ordinal classification settings use expectation. The comparison is shown in \Cref{tb_cls_reg}.

\begin{table}[h]

	\caption{Comparison between different localization methods. }

	\vspace{-1em}
	\setlength{\tabcolsep}{2.5mm}{
	\begin{tabular}{ccccc}
	\toprule
	Dataset     & Regression     & Classification & Ordinal classification \\ \midrule
	CULane & 64.29   &  67.24   & 74.70   \\ 
	\bottomrule
	
    \end{tabular}}
	
	\label{tb_cls_reg}

	\medskip
	\emph{\footnotesize Regression means using the regression as the localization method. Classification is to use the standard classification pipeline to detect lanes. Ordinal classification is associated with the proposed method in this work.}
\end{table}

We can see that the classification with the expectation could gain better performance than the standard classification method. Meanwhile, classification-based methods could consistently outperform the regression-based methods.

\subsubsection{Ablation of the Ordinal Classification Losses}

As described in \Cref{sec_ordinal_cls}, we model the lane detection problem as an ordinal classification problem. To verify the effectiveness of the proposed modules, we show the ablation study of the ordinal classification losses. 

\red{As shown in \Cref{tb_ordinal_cls}, our proposed constraints of expectation loss improves the performance for lane detection effectively. The proposed expectation loss has a different geometric property compared with the standard cross entropy loss, that the expectation loss is like gradually
pushing the prediction's expectation towards the ground truth by reducing the values of logits that are far away from the ground truth and increasing the values of logits that
are near the ground truth. }Meanwhile, the mean average localization error is also reduced compared with the conventional classification.

\begin{table}[h]

	\caption{Ablation study of ordinal classification losses on CULane benchmark with a ResNet-18 backbone.}
	\vspace{-1em}
	\setlength{\tabcolsep}{9mm}{
		\begin{tabular}{ccc}
			\toprule
			Base loss & Exp loss  & F1 (\%)  \\ \midrule
			\checkmark &  & 74.18  \\ 
			 & \checkmark & 69.70 \\
			\checkmark & \checkmark & 74.70   \\
		\bottomrule
	\end{tabular}}
	\label{tb_ordinal_cls}

	\medskip
	\emph{\footnotesize Base loss is the cross-entropy loss. Exp loss is the expectation loss.}
\end{table}

\subsubsection{Effects of the Classification Dimensions}
\label{ab_griding}
\begin{figure}[h]
    \centering

    \centering
    \subfloat[Column Anchor]{\includegraphics[width=0.9\linewidth]{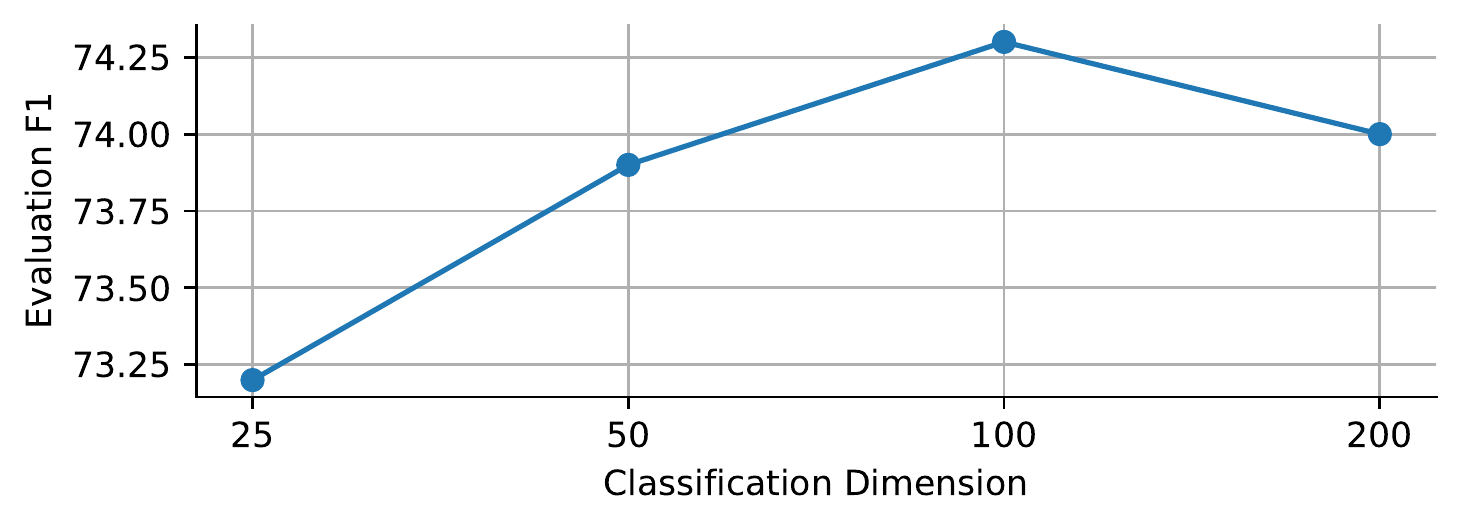}%
    \label{fig_col_dim}}

	\subfloat[Row Anchor]{\includegraphics[width=0.9\linewidth]{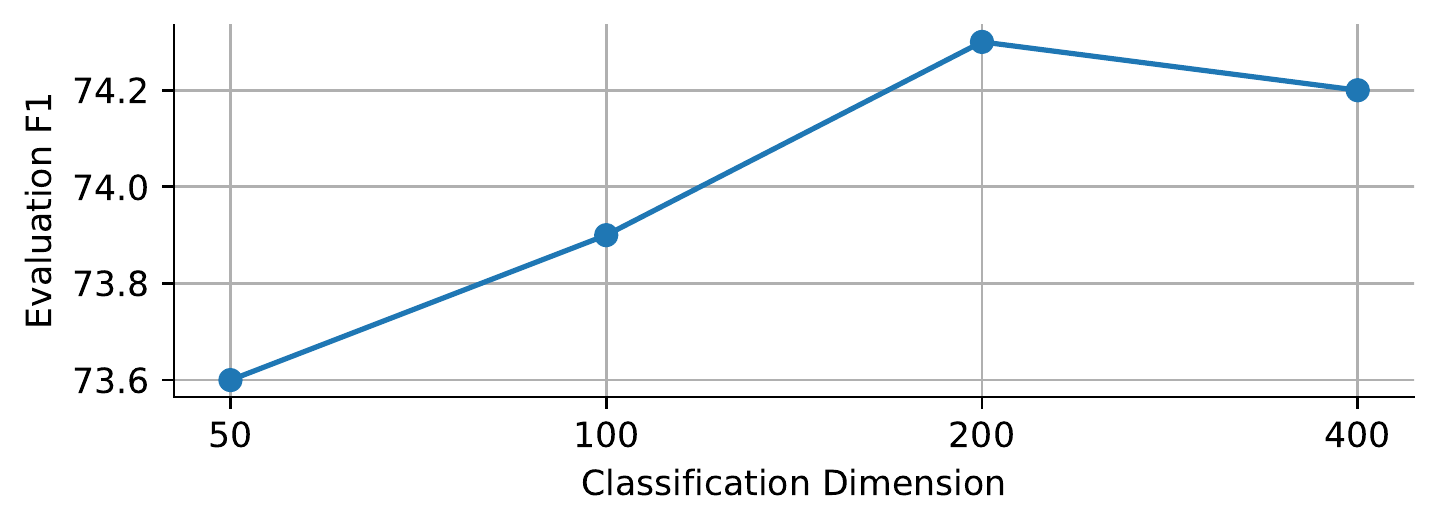}%
    \label{fig_row_dim}}

    \caption{Effects of different classification dimensions on row and column anchors.}
    \label{fig_dim}
\end{figure}

As described in \Cref{eq_net_run}, we use classification-based formulation to detect lanes, and different lane locations are represented with different classes. This raises the question, how many classes are needed to conduct lane detection. To discuss this question, we first set the classification dimension of row anchors $N^r_{dim}$ as 200 and conduct experiments with the classification dimension of column anchors $N^c_{dim}$ as 25, 50, 100, and 200. The results are shown in \Cref{fig_col_dim}. Then, we fix $N^c_{dim}$ as 100, and conduct experiments as $N^r_{dim}$ as 50, 100, 200, and 400. The results are shown in \Cref{fig_row_dim}.

We can see that with the increase of classification dimensions, the performance first increases and then decreases. With a smaller dimension, the classification is inherently easier, but each class represents a wider range of locations, i.e. the localization ability of each class is worse. With a larger dimension, each class represents a narrower range of locations (better localization ability of each class), but the classification itself is harder. The final performance is a trade-off between the difficulty of classification and the localization ability of each class. So we set $N^c_{dim}$ to 100 and $N^r_{dim}$ to 200.

\begin{table*}[t]
    \caption{\red{Comparison of F1-measure and runtime on CULane testing set with IoU threshold=0.5. }}
	\vspace{-1em}
	\begin{tabular}{lccccccccccc}
		\toprule

		\textbf{Method} & \textbf{Normal} & \textbf{Crowded} & \textbf{Dazzle} & \textbf{Shadow} & \textbf{No line} & \textbf{Arrow} & \textbf{Curve} & \textbf{Crossroad} & \textbf{Night} & \textbf{FPS} & \textbf{Total} \\ \midrule

		SCNN~\cite{SCNN} & 90.6  & 69.7   & 58.5        & 66.9  & 43.4   & 84.1 & 64.4 & 1990   & 66.1 & 8                                     & 71.6 \\

		SAD~\cite{SAD} & 90.1  & 68.8   & 60.2        & 65.9  & 41.6   & 84.0 & 65.7 & 1998   & 66.0 & 75                                    & 70.8 \\

		ERFNet-IntRA-KD~\cite{hou2020interregion}   & - & - &   -   & -& - &  -      &     -   &   -     &   -     & 100                                    & 72.4 \\

		SIM-CycleGAN~\cite{SIM}    & 91.8  & 71.8   & 66.4        & 76.2  & 46.1   & 87.8 & 67.1 & 2346   & 69.4 &                -                      & 73.9 \\

		CurveLanes-NAS-S~\cite{xu2020curvelane}    & 88.3  & 68.6   & 63.2        & 68.0  & 47.9   & 82.5 & 66.0 & 2817   & 66.2 &            -                         & 71.4 \\

		CurveLanes-NAS-M~\cite{xu2020curvelane} & 90.2  & 70.5   & 65.9        & 69.3  & 48.8   & 85.7 & 67.5 & 2359   & 68.2 &               -                        & 73.5 \\

		CurveLanes-NAS-L~\cite{xu2020curvelane}  & 90.7  & 72.3   & 67.7        & 70.1  & 49.4   & 85.8 & 68.4 & 1746   & 68.9 &              -                      & 74.8 \\
		RESA (ResNet-34) \cite{zheng2020resa} & 91.9 & 72.4 & 66.5 & 72.0 & 46.3  & 88.1 & 68.6 & 1896 & 69.8 & 45 &  74.5 \\
		RESA (ResNet-50) \cite{zheng2020resa} & 92.1 & 73.1 & 69.2 & 72.8  & 47.7 & 88.3 & \textbf{70.3} & 1503 & 69.9 & 36 &  75.3 \\

		LaneATT (ResNet-18)~\cite{tabelini2021cvpr}    & 91.1  & 72.9   & 65.7        & 70.9  & 48.3   & 85.4 & 63.3 & \textbf{1170}   & 68.9 & 250                                    & 75.1 \\
        
		LaneATT (ResNet-34)~\cite{tabelini2021cvpr}    & 92.1  & 75.0   & 66.4        & 78.1  & 49.3   & 88.3 & 67.7 & 1330   & 70.7 & 171            & 76.6 \\
        
		SGNet~\cite{su2021structure} & 91.4 & 74.0 & 66.8 & 72.1 & 50.1 & 87.1 & 67.0 & 1164 & 70.6 & 117  & 76.1 \\
        
        FOLOLane \cite{qu2021focus} & \textbf{92.7} & \textbf{77.8} & \textbf{75.2} & \textbf{79.3} & \textbf{52.1} & \textbf{89.0} & 69.4  & 1569 & \textbf{74.5}  & 40 & \textbf{78.8} \\
		\midrule
		UFLD(ResNet-18) \cite{qin2020ultra}    & 87.7  & 66.0   & 58.4        & 62.8  & 40.2   & 81.0 & 57.9 & 1743   & 62.1 & 323                        & 68.4 \\

		UFLD(ResNet-34) \cite{qin2020ultra}   & 90.7  & 70.2   & 59.5        & 69.3  & 44.4   & 85.7 & 69.5 & 2037   & 66.7 & 175    & 72.3 \\      
		\midrule
		UFLDv2(ResNet-18)    & 91.7  & 73.0   & 64.6        & 74.7  & 47.2   & 87.6 & 68.7 & 1998   & 70.2 & \textbf{330}                                   & 74.7  \\
		UFLDv2(ResNet-18)*    & 91.8  & 73.3   & 65.3        & 75.1  & 47.6   & 87.9 & 68.5 & 2075   & 70.7 & 310                                   & 75.0 \\

		UFLDv2(ResNet-34)    & 92.5  & 74.9   & 65.7        & 75.3  & 49.0   & 88.5 & 70.2 & 1864   & 70.6 & 165                                    & 75.9  \\
		UFLDv2(ResNet-34)*    & 92.5  & 74.8   & 65.5        & 75.5  & 49.2   & 88.8 & 70.1 & 1910   & 70.8 & 156                                    & 76.0 \\
		\bottomrule
		\end{tabular}
    
	\medskip
	\emph{For crossroads, only false positives are shown. The less, the better.
	`-' means the result is not available. \\
	* means our method with the feature level test time augmentation.}
	\label{tab:CULane compare}

\end{table*}




\subsubsection{Effects of the Number of Anchors}
\label{ab_num_anchor}
\begin{figure}[h]
    \centering

    \centering
	\subfloat[Column Anchor]{\includegraphics[width=0.9\linewidth]{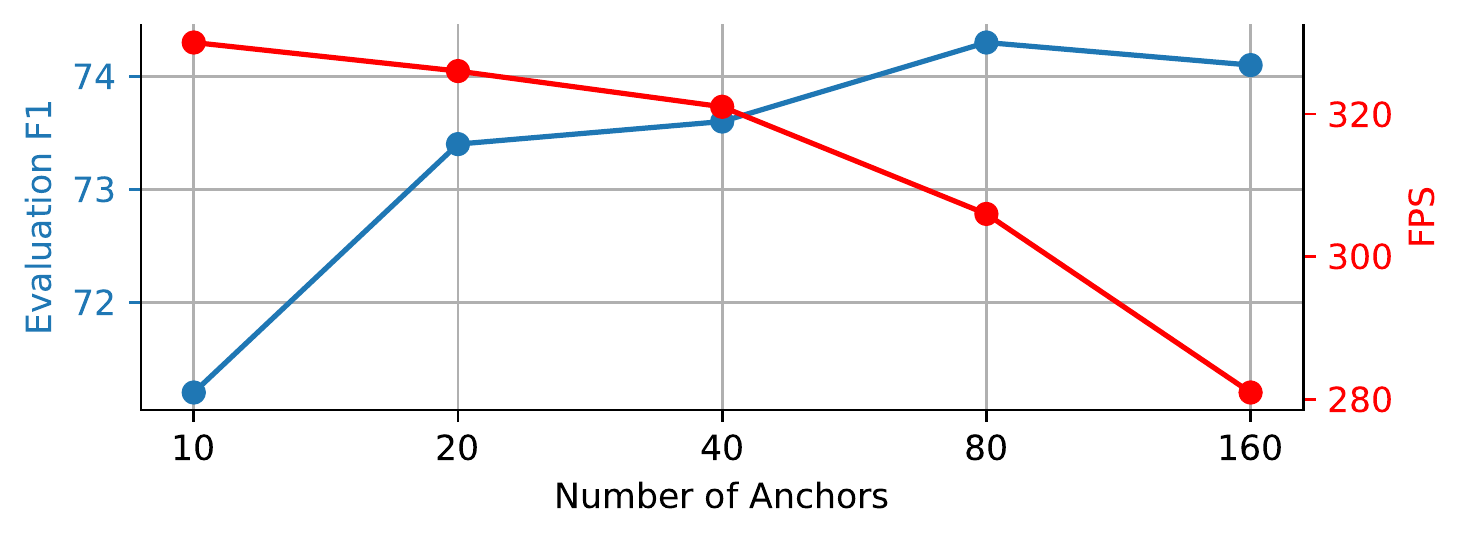}%
    \label{fig_col_num_anchor}}

    \subfloat[\red{Row Anchor}]{\includegraphics[width=0.9\linewidth]{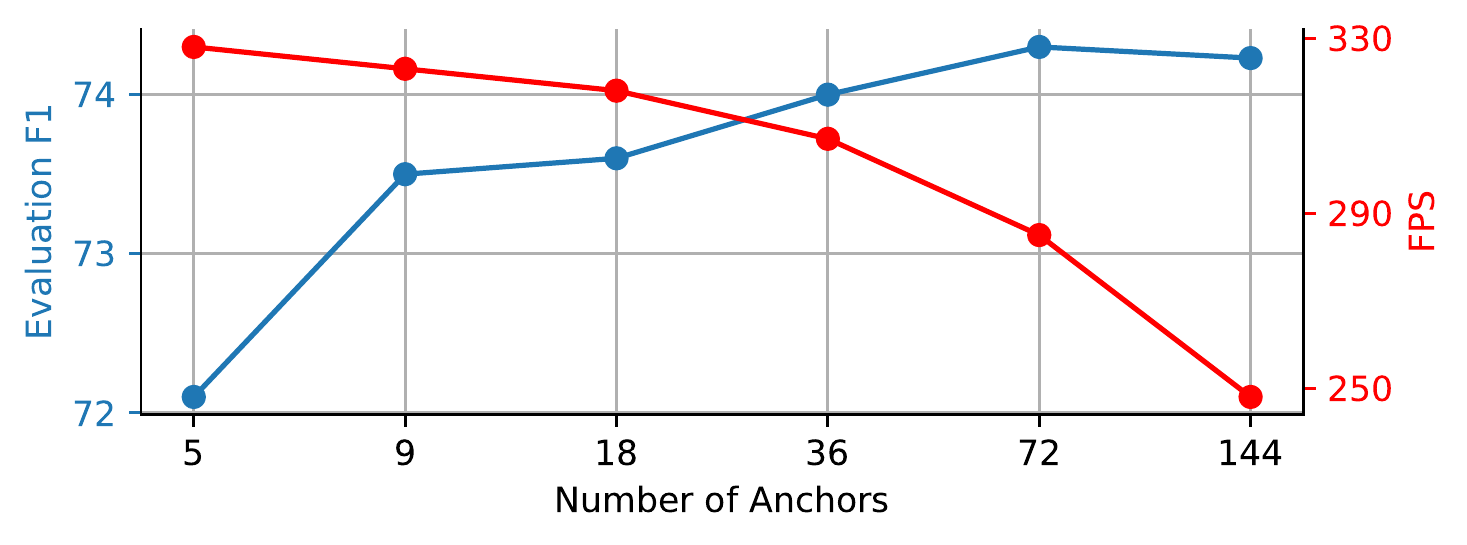}%
    \label{fig_row_num_anchor}}

    \caption{\red{Effects of using different number of row and column anchors.}}
    \label{fig_num_anchor}
\end{figure}

\red{There is another important hyperparameter in our method: the number of anchors ($N_{row}$ and $N_{col}$) used to represent lanes.} In this way, we set the number of row anchors $N_{row}$ to 18 (following \cite{SCNN}) and conduct experiments with the number of column anchors $N_{col}$ as 10, 20, 40, 80, and 160. The results are shown in \Cref{fig_col_num_anchor}. Then, we set $N_{col}$ to 40 and conduct experiments with $N_{row}$ as 5, 9, 18, 36, and 72. The results are shown in \Cref{fig_row_num_anchor}.

We can see that with the increase of the number of row anchors, the performance also increases generally. But the detection speed also drops gradually. In this way, we set $N_{row}$ and $N_{col}$ to 18 and 40, respectively, to get a balance between performance and speed for ResNet-18-based models. For large models like ResNet-34, we set $N_{row}$ and $N_{col}$ to 72 and 80, respectively.

\subsection{Results}

\begin{table}[hbt]
	\caption{\red{Comparison with other methods on TuSimple test set.} }
	\label{tab_tusimple}
	\vspace{-1em}
    \setlength{\tabcolsep}{0.6mm}{
	\begin{tabular}{lcccccc}
		\toprule
		Method   & F1 & Accuracy  & FP & FN & Runtime & Speed up \\ \midrule
		Res18-Seg\cite{chen2017deeplab} & & 92.69 &&   & 25.3        &   5.3x       \\
		Res34-Seg\cite{chen2017deeplab}  & & 92.84  &&  & 50.5        &   2.6x      \\
		LaneNet\cite{End-to-End}   & 94.80 & 96.38  & 7.80 &  2.44 & 19.0        &   7.0x      \\
		EL-GAN \cite{ghafoorian2018gan}  & 96.26 & 94.90 & 4.12 &  3.36 & $>$100        &   $<$1.3x   \\
		SCNN \cite{SCNN}    & 95.97 & 96.53  & 6.17 & \textbf{1.80} & 133.5       &   1.0x       \\
		SAD  \cite{SAD}   & 95.92  & \textbf{96.64} & 6.02 &  2.05  & 13.4        &   10.0x     \\
	    FastDraw \cite{FastDraw} & 93.92 & 95.20 & 7.60 & 5.40   & 11.1        &   12.0x     \\
	    LSTR \cite{LSTR} & \textbf{96.85} & 96.18 & \textbf{2.91} & 3.38   & 20.8        &   6.3x     \\
		SGNet \cite{su2021structure} & & 95.87 && & 10.9 & 12.3x\\
		LaneATT (Res18) \cite{tabelini2021cvpr} & 96.71 & 95.57 &3.56& 3.01 & 4.0 & 33.4x \\
		LaneATT (Res34) \cite{tabelini2021cvpr} & 96.77 & 95.63 & 3.53 & 2.92 & 5.9 & 22.6x \\
		\midrule 
		\textit{protocol 1} \\
		UFLD(ResNet-18)     & 87.87 & 95.82  & 19.05 & 3.92 & 3.2         &   41.7x     \\ 
		UFLD(ResNet-34)     & 87.91 & 95.81   & 19.02 & 3.85 & 5.9        &   22.6x     \\
		UFLDv2(ResNet-18)   & 87.90 & 95.84 & 18.98 & 3.94  & 3.2         &   41.7x     \\ 
		UFLDv2(ResNet-34)   & 88.08 & 95.73 & 18.84 & 3.70  & 5.9         &   22.6x     \\ 
		\midrule
		
		\textit{protocol 2} \\
		UFLD(ResNet-18)     & 96.05 & 95.50  & 3.06 & 4.82 & 3.2         &   41.7x     \\ 
		UFLD(ResNet-34)   & 96.13 & 95.53 & 3.06 & 4.66  & 5.9         &   22.6x     \\ 
		UFLDv2(ResNet-18)   & 96.16 & 95.65 & 3.06 & 4.61  & 3.2         &   41.7x     \\ 
		UFLDv2(ResNet-34)   & 96.22 & 95.56 & 3.18 & 4.37  & 5.9         &   22.6x     \\ 
		\bottomrule
	\end{tabular}}

    \medskip
	\emph{\footnotesize The calculation of speed up is based on the slowest method SCNN. \\
    \footnotesize The speed of LSTR is measured on Nvidia RTX 3090 GPU with CUDA 11.1, PyTorch 1.8.2, and batchsize=1.}
\end{table}

\red{This section shows the results on the four lane detection datasets, which are the TuSimple, CULane, CurveLanes, and LLAMAS datasetS.} In these experiments, ResNet-18 and ResNet-34 are used as our backbone models.

\red{For the CULane dataset, seven methods, including SCNN \cite{SCNN}, SAD \cite{SAD}, ERFNet~\cite{hou2020interregion}, SIM-CycleGAN~\cite{SIM}, CurveLanes~\cite{xu2020curvelane}, SGNet~\cite{su2021structure}, RESA~\cite{zheng2020resa}, LaneATT~\cite{tabelini2021cvpr} and FOLOLane~\cite{qu2021focus}, are used for comparison.} F1-measure and runtime are compared. The results can be seen in \Cref{tab:CULane compare}.

It is observed from \Cref{tab:CULane compare} that our method achieves the fastest speed. Compared with the previous conference version, we can get stronger results with 6.3 points of performance improvement at the same speed. It proves the effectiveness of the proposed formulation on these challenging scenarios because our method could utilize global information to address the no-visual-clue and efficiency problems. The fastest model achieves 300+ FPS.

\red{For the TuSimple lane detection benchmark, ten methods are used for comparison, including Res18-Seg \cite{chen2017deeplab}, Res34-Seg \cite{chen2017deeplab}, LaneNet \cite{End-to-End}, EL-GAN \cite{ghafoorian2018gan}, SCNN \cite{SCNN}, SAD \cite{SAD}, SGNet~\cite{su2021structure}, LaneATT~\cite{tabelini2021cvpr}, and LSTR~\cite{LSTR}. F1, Accuracy, FP, FN and runtime are compared in this experiment. The runtime of our method is recorded with the average time for 100 runs. To validate our methods, we use two protocols. Protocol 1 outputs all lanes and the absent lanse are represented with lanes full of \texttt{invalid}, which is the same as the conference version. Protocol 2 directly discards the absent lanes. The results are shown in \Cref{tab_tusimple}.}

From \Cref{tab_tusimple}, we can see that our method achieves comparable performance with state-of-the-art methods while our method could run extremely fast. The biggest runtime gap between our method and SCNN is that our method could infer 41.7 times faster.

\red{Another interesting phenomenon we should notice is that our method gains both better performance and faster speed when the backbone network is the same as plain segmentation, which is a DeeplabV2 \cite{chen2017deeplab} model with the resnet backbones.} This result shows that our method is better than the segmentation formulation and verifies the effectiveness of our formulation.

\begin{table}[h]
    \centering
    \caption{\red{Comparison on the CurveLane dataset.}}
    \label{tb_curvelane}
    \vspace{-1em}
    \setlength{\tabcolsep}{3.5mm}{
    \begin{tabular}{lcccc}
    \toprule
    Method       & F1    & Presicion & Recall & FPS \\ \midrule
    SCNN \cite{SCNN}        & 65.02 & 76.13     & 56.74  & 8   \\
    SAD  \cite{SAD}       & 50.31 & 63.60     & 41.60  & 75  \\
    PointLaneNet \cite{chen2019pointlanenet} & 78.47 & 86.33     & 72.91  & 111 \\
    CurveLane-S \cite{xu2020curvelane} & 81.12 & \textbf{93.58}     & 71.59  & \textless45  \\
    UFLDv2(ResNet-18)  & 80.45 & 81.49     & 79.44  &   \textbf{145}  \\ 
    UFLDv2(ResNet-34)  & \textbf{81.34} & 81.93     & \textbf{80.76}  &   86  \\ 
    
    \bottomrule
    \end{tabular}}
    \end{table}
\red{For the CurveLanes dataset, we show the results in \Cref{tb_curvelane}. We can see that our method achieves a better performance while maintaining a faster speed compared with the CurveLane-S method.}

\begin{table}[h]
    \centering
    \caption{\red{Comparison on the LLAMAS dataset.}}
    \vspace{-1em}
    \label{tab_llamas}
    \begin{tabular}{lcccc}
        \toprule
    Method              & F1    & Precision & Recall & FPS \\
    \midrule
    PolyLaneNet \cite{tabelini2021polylanenet}       & 88.40 & 88.87     & 87.93  & 115 \\
    LaneATT(ResNet-18) \cite{tabelini2021cvpr}  & 93.46 & \textbf{96.92}     & 90.24  & 250 \\
    LaneATT(ResNet-34) \cite{tabelini2021cvpr} & 93.74 & 96.79     & 90.88  & 171 \\
    LaneATT(ResNet-122) \cite{tabelini2021cvpr} & 93.54 & 96.82     & 90.47  & 26  \\
    UFLDv2(ResNet-18)   & 94.58 & 95.29     & 93.88  & \textbf{285} \\
    UFLDv2(ResNet-34)   & \textbf{94.95} & 95.75     & \textbf{94.17}  & 156  \\
    \bottomrule
    \end{tabular}
    \end{table}

\red{For the LLAMAS dataset, we show the results in \Cref{tab_llamas}. It can be seen that our method also achieves the best performance and the fastest speed.}

\begin{figure}[h]
    \centering
    \includegraphics[width = \linewidth]{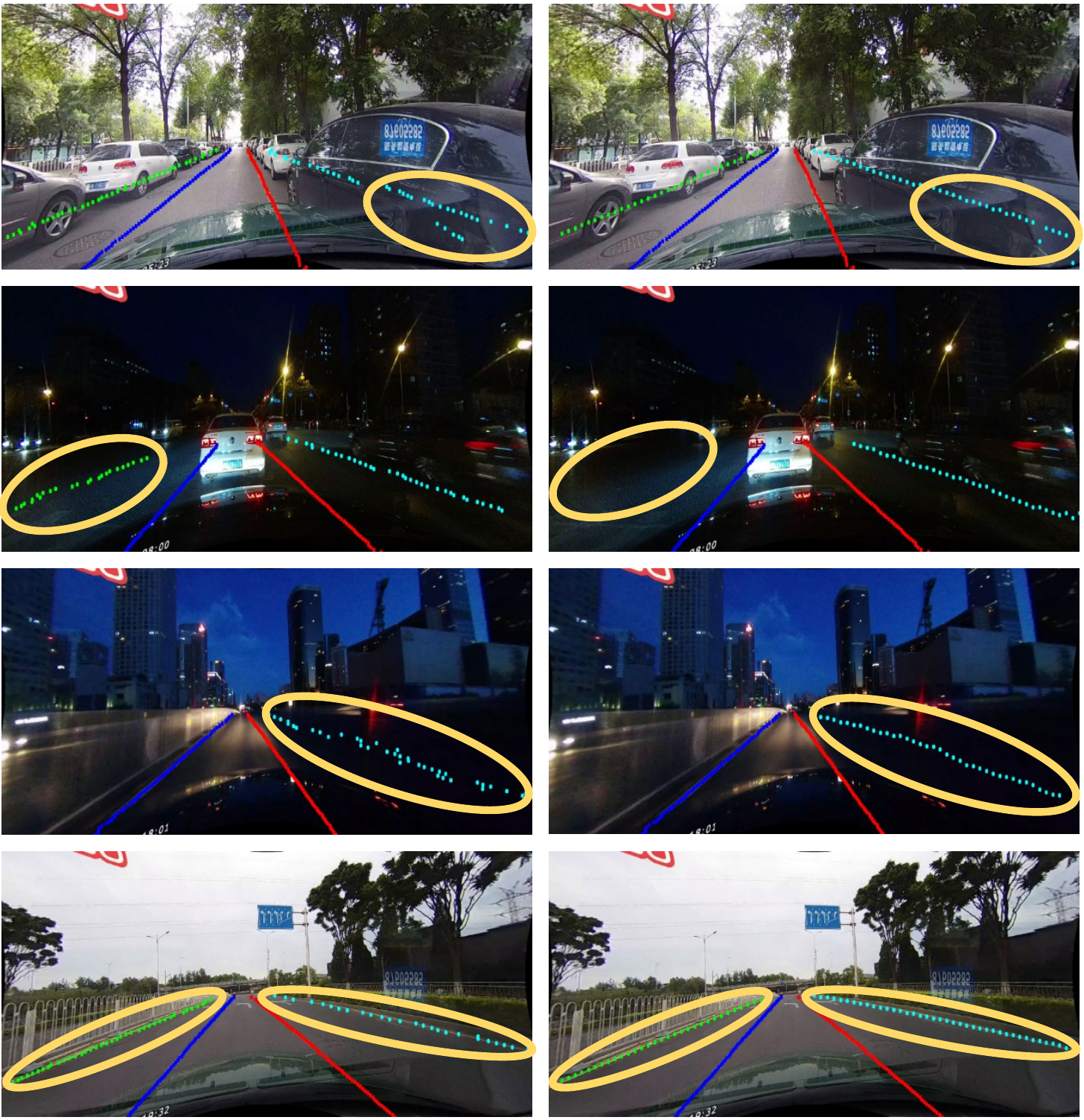}
    \caption{\red{Visual comparisons between row and hybrid anchors  under occluded, night, and normal settings. We show the results of the row-anchor model on the left and the hybrid-anchor model on the right. Yellow ellipses are used to highlight the differences.}}
    \label{fig_vis_comp}
\end{figure}

\begin{figure*}[h]

	\centering
	\includegraphics[width=0.9\linewidth, height=1.\linewidth]{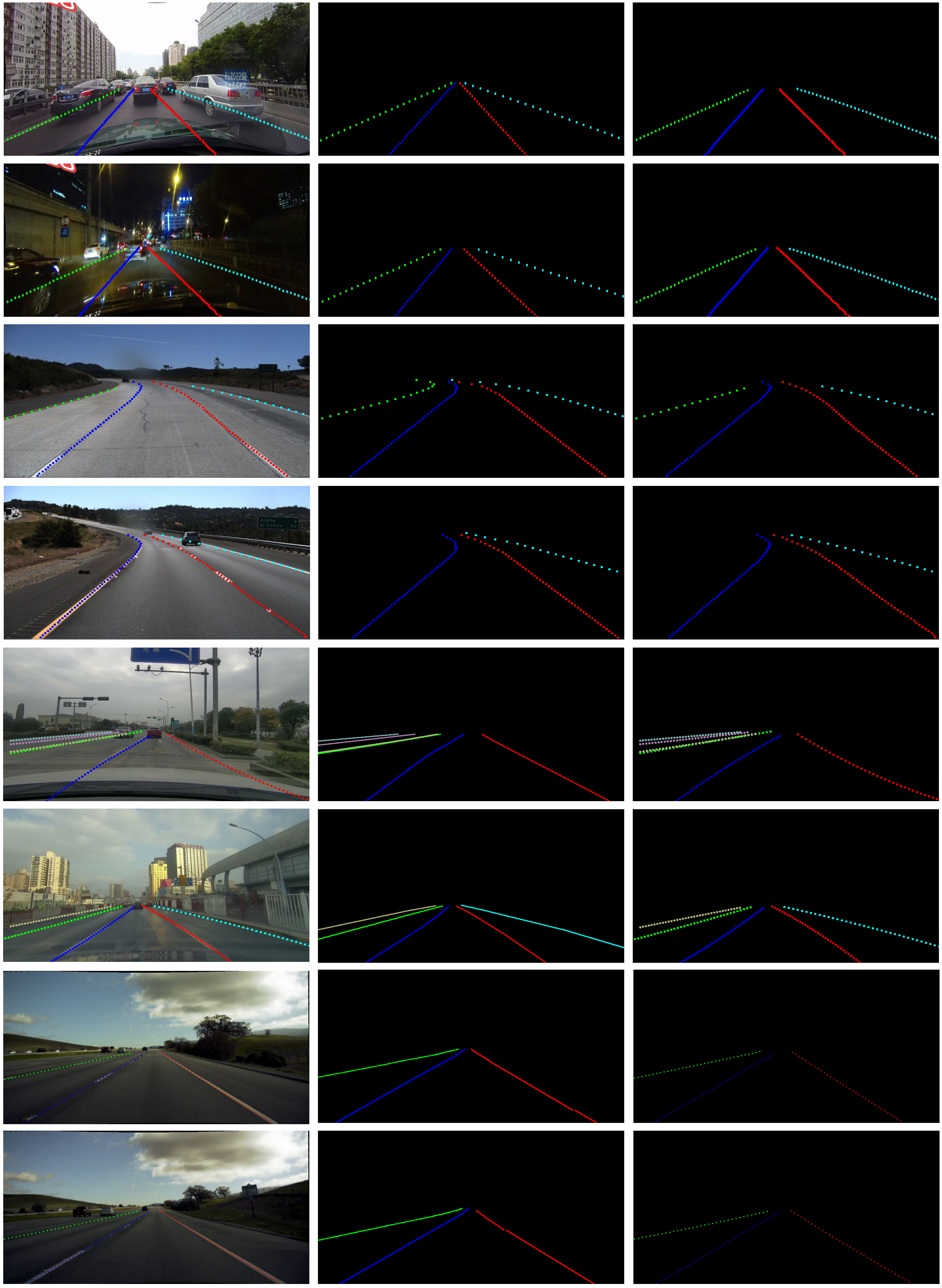}
	\caption{ \red{Visualization on the CULane, TuSimple, CurveLanes, and LLAMAS datasets. Each dataset is visualized with two rows. From left to right, the results are images, ground truths, and predictions, respectively. In the image, predicted lane points are annotated.}}
	\label{fig_vis}
\end{figure*}

\red{The visualizations of our method on the four datasets are shown in \Cref{fig_vis_comp,fig_vis}. We can see that the proposed hybrid anchors and our method perform well under various conditions.} 

\section{Conclusion}
\label{sec_conclusion}

In this paper, we have proposed a novel formulation with a hybrid anchor system and ordinal classification pipeline to achieve both remarkable speed and accuracy. The proposed formulation regards lane detection as directly learning the sparse coordinates on a hybrid anchor system with top-down classification based on the global feature. In this way, the problems of efficiency and no-visual-clue can be effectively addressed. The effectiveness of the proposed hybrid anchor system and ordinal classification losses are justified with both qualitative and quantitative experiments. Especially, a lightweight ResNet-18 version of our method could even achieve 300+ FPS.

\red{There also exists some weakness of our method that the arrangements of anchors, i.e., how many anchors are to be set and where to place anchors, are still fixed and handcrafted. Automatic, dynamic, rotatable, and nonuniform anchors would be the potential directions for stronger and even faster methods. }
\vspace{-5pt}
\ifCLASSOPTIONcompsoc
  \section*{Acknowledgments}
\else
  \section*{Acknowledgment}
\fi

This work was supported in part by National Key Research and Development Program of China under Grant 2020AAA0107400, Zhejiang Provincial Natural Science Foundation of China under Grant LR19F020004, National Natural Science Foundation of China under Grant U20A20222, and OPPO Research Fund. The authors would like to thank Huanyu Wang for his contribution in idea discussion.

\ifCLASSOPTIONcaptionsoff
  \newpage
\fi

\bibliographystyle{IEEEtran}
\bibliography{egbib}

\begin{thebibliography}{10}
\providecommand{\url}[1]{#1}
\csname url@samestyle\endcsname
\providecommand{\newblock}{\relax}
\providecommand{\bibinfo}[2]{#2}
\providecommand{\BIBentrySTDinterwordspacing}{\spaceskip=0pt\relax}
\providecommand{\BIBentryALTinterwordstretchfactor}{4}
\providecommand{\BIBentryALTinterwordspacing}{\spaceskip=\fontdimen2\font plus
\BIBentryALTinterwordstretchfactor\fontdimen3\font minus
  \fontdimen4\font\relax}
\providecommand{\BIBforeignlanguage}[2]{{%
\expandafter\ifx\csname l@#1\endcsname\relax
\typeout{** WARNING: IEEEtran.bst: No hyphenation pattern has been}%
\typeout{** loaded for the language `#1'. Using the pattern for}%
\typeout{** the default language instead.}%
\else
\language=\csname l@#1\endcsname
\fi
#2}}
\providecommand{\BIBdecl}{\relax}
\BIBdecl

\bibitem{hillel2014recent}
A.~B. Hillel, R.~Lerner, D.~Levi, and G.~Raz, ``Recent progress in road and
  lane detection: a survey,'' \emph{Mach. Vis. Appl.}, vol.~25, no.~3, pp.
  727--745, 2014.

\bibitem{SCNN}
X.~Pan, J.~Shi, P.~Luo, X.~Wang, and X.~Tang, ``Spatial as deep: Spatial cnn
  for traffic scene understanding,'' in \emph{AAAI}, 2018, pp. 7276--7284.

\bibitem{SAD}
Y.~Hou, Z.~Ma, C.~Liu, and C.~C. Loy, ``Learning lightweight lane detection
  cnns by self attention distillation,'' in \emph{Int. Conf. Comput. Vis.},
  2019, pp. 1013--1021.

\bibitem{Lee_2017_ICCV}
S.~Lee, J.~Kim, J.~Shin~Yoon, S.~Shin, O.~Bailo, N.~Kim, T.-H. Lee,
  H.~Seok~Hong, S.-H. Han, and I.~So~Kweon, ``Vpgnet: Vanishing point guided
  network for lane and road marking detection and recognition,'' in \emph{Int.
  Conf. Comput. Vis.}, Oct 2017, pp. 1947--1955.

\bibitem{End-to-End}
D.~Neven, B.~De~Brabandere, S.~Georgoulis, M.~Proesmans, and L.~Van~Gool,
  ``Towards end-to-end lane detection: an instance segmentation approach,'' in
  \emph{IEEE Intell. Veh. Symp.}, 2018, pp. 286--291.

\bibitem{xu2020curvelane}
H.~Xu, S.~Wang, X.~Cai, W.~Zhang, X.~Liang, and Z.~Li, ``Curvelane-nas:
  Unifying lane-sensitive architecture search and adaptive point blending,'' in
  \emph{Eur. Conf. Comput. Vis.}, 2020, pp. 1--16.

\bibitem{7410526}
R.~Girshick, ``Fast r-cnn,'' in \emph{Int. Conf. Comput. Vis.}, 2015, pp.
  1440--1448.

\bibitem{7485869}
S.~Ren, K.~He, R.~Girshick, and J.~Sun, ``Faster r-cnn: Towards real-time
  object detection with region proposal networks,'' \emph{IEEE Trans. Pattern
  Anal. Mach. Intell.}, vol.~39, no.~6, pp. 1137--1149, 2017.

\bibitem{redmon2016you}
J.~Redmon, S.~Divvala, R.~Girshick, and A.~Farhadi, ``You only look once:
  Unified, real-time object detection,'' in \emph{IEEE Conf. Comput. Vis.
  Pattern Recog.}, 2016, pp. 779--788.

\bibitem{liu2016ssd}
W.~Liu, D.~Anguelov, D.~Erhan, C.~Szegedy, S.~Reed, C.-Y. Fu, and A.~C. Berg,
  ``Ssd: Single shot multibox detector,'' in \emph{Eur. Conf. Comput.
  Vis.}\hskip 1em plus 0.5em minus 0.4em\relax Springer, 2016, pp. 21--37.

\bibitem{tian2021fcos}
Z.~Tian, C.~Shen, H.~Chen, and T.~He, ``Fcos: A simple and strong anchor-free
  object detector,'' \emph{IEEE Trans. Pattern Anal. Mach. Intell.}, vol.~1,
  no.~1, pp. 1--12, 2020.

\bibitem{6619290}
Y.~Sun, X.~Wang, and X.~Tang, ``Deep convolutional network cascade for facial
  point detection,'' in \emph{IEEE Conf. Comput. Vis. Pattern Recog.}, 2013,
  pp. 3476--3483.

\bibitem{zhang2014facial}
Z.~Zhang, P.~Luo, C.~C. Loy, and X.~Tang, ``Facial landmark detection by deep
  multi-task learning,'' in \emph{Eur. Conf. Comput. Vis.}\hskip 1em plus 0.5em
  minus 0.4em\relax Springer, 2014, pp. 94--108.

\bibitem{7553523}
K.~Zhang, Z.~Zhang, Z.~Li, and Y.~Qiao, ``Joint face detection and alignment
  using multitask cascaded convolutional networks,'' \emph{IEEE Sign. Process.
  Letters}, vol.~23, no.~10, pp. 1499--1503, 2016.

\bibitem{8765346}
Z.~Cao, G.~Hidalgo, T.~Simon, S.-E. Wei, and Y.~Sheikh, ``Openpose: Realtime
  multi-person 2d pose estimation using part affinity fields,'' \emph{IEEE
  Trans. Pattern Anal. Mach. Intell.}, vol.~43, no.~1, pp. 172--186, 2021.

\bibitem{wei2016convolutional}
S.-E. Wei, V.~Ramakrishna, T.~Kanade, and Y.~Sheikh, ``Convolutional pose
  machines,'' in \emph{IEEE Conf. Comput. Vis. Pattern Recog.}, 2016, pp.
  4724--4732.

\bibitem{newell2016stacked}
A.~Newell, K.~Yang, and J.~Deng, ``Stacked hourglass networks for human pose
  estimation,'' in \emph{Eur. Conf. Comput. Vis.}\hskip 1em plus 0.5em minus
  0.4em\relax Springer, 2016, pp. 483--499.

\bibitem{deng2009imagenet}
J.~Deng, W.~Dong, R.~Socher, L.-J. Li, K.~Li, and L.~Fei-Fei, ``Imagenet: A
  large-scale hierarchical image database,'' in \emph{IEEE Conf. Comput. Vis.
  Pattern Recog.}\hskip 1em plus 0.5em minus 0.4em\relax Ieee, 2009, pp.
  248--255.

\bibitem{7406390}
R.~Rothe, R.~Timofte, and L.~Van~Gool, ``Dex: Deep expectation of apparent age
  from a single image,'' in \emph{IEEE Conf. Comput. Vis. Worksh.}, 2015, pp.
  252--257.

\bibitem{qin2020ultra}
Z.~Qin, H.~Wang, and X.~Li, ``Ultra fast structure-aware deep lane detection,''
  in \emph{Eur. Conf. Comput. Vis.}, 2020, pp. 1--16.

\bibitem{sun2006hsi}
T.-Y. Sun, S.-J. Tsai, and V.~Chan, ``Hsi color model based lane-marking
  detection,'' in \emph{IEEE Trans. Intell. Transp. Syst. Conf.}, 2006, pp.
  1168--1172.

\bibitem{yu1997lane}
B.~Yu and A.~K. Jain, ``Lane boundary detection using a multiresolution hough
  transform,'' in \emph{IEEE Int. Conf. Image Process.}, vol.~2, 1997, pp.
  748--751.

\bibitem{wang2000lane}
Y.~Wang, D.~Shen, and E.~K. Teoh, ``Lane detection using spline model,''
  \emph{Pattern Recognit. Lett.}, vol.~21, no.~8, pp. 677--689, 2000.

\bibitem{bertozzi1998gold}
M.~Bertozzi and A.~Broggi, ``Gold: A parallel real-time stereo vision system
  for generic obstacle and lane detection,'' \emph{IEEE Trans. Image Process.},
  vol.~7, no.~1, pp. 62--81, 1998.

\bibitem{aly2008real}
M.~Aly, ``Real time detection of lane markers in urban streets,'' in \emph{IEEE
  Intell. Veh. Symp.}, 2008, pp. 7--12.

\bibitem{wang2004lane}
Y.~Wang, E.~K. Teoh, and D.~Shen, ``Lane detection and tracking using
  b-snake,'' \emph{Image Vis. Comput.}, vol.~22, no.~4, pp. 269--280, 2004.

\bibitem{kim2008robust}
Z.~Kim, ``Robust lane detection and tracking in challenging scenarios,''
  \emph{IEEE Trans. Intell. Transp. Syst.}, vol.~9, no.~1, pp. 16--26, 2008.

\bibitem{CRF}
P.~Kr{\"a}henb{\"u}hl and V.~Koltun, ``Efficient inference in fully connected
  crfs with gaussian edge potentials,'' in \emph{Adv. Neural Inform. Process.
  Syst.}, 2011, pp. 109--117.

\bibitem{kluge1995deformable}
K.~Kluge and S.~Lakshmanan, ``A deformable-template approach to lane
  detection,'' in \emph{IEEE Intell. Veh. Symp.}, 1995, pp. 54--59.

\bibitem{gonzalez2000lane}
J.~P. Gonzalez and U.~Ozguner, ``Lane detection using histogram-based
  segmentation and decision trees,'' in \emph{IEEE Trans. Intell. Transp. Syst.
  Conf.}, 2000, pp. 346--351.

\bibitem{mandalia2005using}
H.~M. Mandalia and M.~D.~D. Salvucci, ``Using support vector machines for
  lane-change detection,'' in \emph{Proc. Hum. Factors Ergon. Soc. Annu.
  Meet.}, vol.~49, no.~22, 2005, pp. 1965--1969.

\bibitem{kim2014robust}
J.~Kim and M.~Lee, ``Robust lane detection based on convolutional neural
  network and random sample consensus,'' in \emph{ICONIP}, 2014, pp. 454--461.

\bibitem{Empirical_Evaluation}
B.~Huval, T.~Wang, S.~Tandon, J.~Kiske, W.~Song, J.~Pazhayampallil,
  M.~Andriluka, P.~Rajpurkar, T.~Migimatsu, R.~Cheng-Yue \emph{et~al.}, ``An
  empirical evaluation of deep learning on highway driving,'' \emph{arXiv
  preprint arXiv:1504.01716}, 2015.

\bibitem{chng2020roneld}
Z.~M. Chng, J.~M.~H. Lew, and J.~A. Lee, ``Roneld: Robust neural network output
  enhancement for active lane detection,'' 2020.

\bibitem{zheng2020resa}
T.~Zheng, H.~Fang, Y.~Zhang, W.~Tang, Z.~Yang, H.~Liu, and D.~Cai, ``Resa:
  Recurrent feature-shift aggregator for lane detection,'' 2020.

\bibitem{hou2020interregion}
Y.~Hou, Z.~Ma, C.~Liu, T.-W. Hui, and C.~C. Loy, ``Inter-region affinity
  distillation for road marking segmentation,'' in \emph{IEEE Conf. Comput.
  Vis. Pattern Recog.}, June 2020.

\bibitem{abualsaud2021laneaf}
H.~Abualsaud, S.~Liu, D.~Lu, K.~Situ, A.~Rangesh, and M.~M. Trivedi, ``Laneaf:
  Robust multi-lane detection with affinity fields,'' \emph{arXiv preprint
  arXiv:2103.12040}, 2021.

\bibitem{qu2021focus}
Z.~Qu, H.~Jin, Y.~Zhou, Z.~Yang, and W.~Zhang, ``Focus on local: Detecting lane
  marker from bottom up via key point,'' \emph{arXiv preprint
  arXiv:2105.13680}, 2021.

\bibitem{li2016deep}
J.~Li, X.~Mei, D.~Prokhorov, and D.~Tao, ``Deep neural network for structural
  prediction and lane detection in traffic scene,'' \emph{IEEE Trans. Neural
  Netw. Learn. Syst.}, vol.~28, no.~3, pp. 690--703, 2016.

\bibitem{FastDraw}
J.~Philion, ``Fastdraw: Addressing the long tail of lane detection by adapting
  a sequential prediction network,'' in \emph{IEEE Conf. Comput. Vis. Pattern
  Recog.}, 2019, pp. 11\,582--11\,591.

\bibitem{Proposal-Free}
Y.-C. Hsu, Z.~Xu, Z.~Kira, and J.~Huang, ``Learning to cluster for
  proposal-free instance segmentation,'' in \emph{IJCNN}, 2018, pp. 1--8.

\bibitem{wvangansbeke_2019}
W.~Van~Gansbeke, B.~De~Brabandere, D.~Neven, M.~Proesmans, and L.~Van~Gool,
  ``End-to-end lane detection through differentiable least-squares fitting,''
  \emph{arXiv preprint arXiv:1902.00293}, 2019.

\bibitem{tabelini2021polylanenet}
L.~Tabelini, R.~Berriel, T.~M. Paixao, C.~Badue, A.~F. De~Souza, and
  T.~Oliveira-Santos, ``Polylanenet: Lane estimation via deep polynomial
  regression,'' in \emph{Int. Conf. Pattern Recog.}\hskip 1em plus 0.5em minus
  0.4em\relax IEEE, 2021, pp. 6150--6156.

\bibitem{LSTR}
R.~Liu, Z.~Yuan, T.~Liu, and Z.~Xiong, ``End-to-end lane shape prediction with
  transformers,'' in \emph{WACV}, 2021, pp. 3694--3702.

\bibitem{NIPS2017_3f5ee243}
A.~Vaswani, N.~Shazeer, N.~Parmar, J.~Uszkoreit, L.~Jones, A.~N. Gomez, L.~u.
  Kaiser, and I.~Polosukhin, ``Attention is all you need,'' in \emph{Adv.
  Neural Inform. Process. Syst.}, 2017, pp. 5998--6008.

\bibitem{tabelini2021cvpr}
L.~Tabelini, R.~Berriel, T.~M.~P. ao, C.~Badue, A.~F.~D. Souza, and
  T.~Oliveira-Santos, ``Keep your eyes on the lane: Real-time attention-guided
  lane detection,'' in \emph{IEEE Conf. Comput. Vis. Pattern Recog.}, 2021, pp.
  1--9.

\bibitem{su2021structure}
J.~Su, C.~Chen, K.~Zhang, J.~Luo, X.~Wei, and X.~Wei, ``Structure guided lane
  detection,'' \emph{arXiv preprint arXiv:2105.05403}, 2021.

\bibitem{garnett20193d}
N.~Garnett, R.~Cohen, T.~Pe'er, R.~Lahav, and D.~Levi, ``3d-lanenet: end-to-end
  3d multiple lane detection,'' in \emph{Int. Conf. Comput. Vis.}, 2019, pp.
  2921--2930.

\bibitem{guo2020gen}
Y.~Guo, G.~Chen, P.~Zhao, W.~Zhang, J.~Miao, J.~Wang, and T.~E. Choe,
  ``Gen-lanenet: A generalized and scalable approach for 3d lane detection,''
  pp. 1--16, 2020.

\bibitem{efrat2020semi}
N.~Efrat, M.~Bluvstein, N.~Garnett, D.~Levi, S.~Oron, and B.~E. Shlomo,
  ``Semi-local 3d lane detection and uncertainty estimation,'' \emph{arXiv
  preprint arXiv:2003.05257}, 2020.

\bibitem{sela20203d}
N.~E. Sela, M.~Bluvstein, S.~Oron, D.~Levi, N.~Garnett, and B.~E. Shlomo,
  ``3d-lanenet+: Anchor free lane detection using a semi-local
  representation,'' \emph{arXiv preprint arXiv:2011.01535}, 2020.

\bibitem{liu2021condlanenet}
L.~Liu, X.~Chen, S.~Zhu, and P.~Tan, ``Condlanenet: a top-to-down lane
  detection framework based on conditional convolution,'' \emph{arXiv preprint
  arXiv:2105.05003}, 2021.

\bibitem{tusimple}
TuSimple, ``Tusimple benchmark,''
  \url{https://github.com/TuSimple/tusimple-benchmark}, accessed November,
  2019.

\bibitem{Simonyan15}
K.~Simonyan and A.~Zisserman, ``Very deep convolutional networks for
  large-scale image recognition,'' in \emph{Int. Conf. Learn. Represent.},
  2015, pp. 1--14.

\bibitem{szegedy2015going}
C.~Szegedy, W.~Liu, Y.~Jia, P.~Sermanet, S.~Reed, D.~Anguelov, D.~Erhan,
  V.~Vanhoucke, and A.~Rabinovich, ``Going deeper with convolutions,'' in
  \emph{IEEE Conf. Comput. Vis. Pattern Recog.}, 2015, pp. 1--9.

\bibitem{resnet}
K.~He, X.~Zhang, S.~Ren, and J.~Sun, ``Deep residual learning for image
  recognition,'' in \emph{IEEE Conf. Comput. Vis. Pattern Recog.}, 2016, pp.
  770--778.

\bibitem{xie2017aggregated}
S.~Xie, R.~Girshick, P.~Doll{\'a}r, Z.~Tu, and K.~He, ``Aggregated residual
  transformations for deep neural networks,'' in \emph{IEEE Conf. Comput. Vis.
  Pattern Recog.}, 2017, pp. 1492--1500.

\bibitem{romera2017erfnet}
E.~Romera, J.~M. Alvarez, L.~M. Bergasa, and R.~Arroyo, ``Erfnet: Efficient
  residual factorized convnet for real-time semantic segmentation,'' \emph{IEEE
  Trans. Intell. Transp. Syst.}, vol.~19, no.~1, pp. 263--272, 2017.

\bibitem{llamas2019}
K.~Behrendt and R.~Soussan, ``Unsupervised labeled lane markers using maps,''
  in \emph{IEEE Conf. Comput. Vis. Worksh.}, 2019.

\bibitem{paszke2017automatic}
A.~Paszke, S.~Gross, S.~Chintala, G.~Chanan, E.~Yang, Z.~DeVito, Z.~Lin,
  A.~Desmaison, L.~Antiga, and A.~Lerer, ``Automatic differentiation in
  pytorch,'' in \emph{Adv. Neural Inform. Process. Syst. Worksh.}, 2017, pp.
  1--4.

\bibitem{SIM}
T.~Liu, Z.~Chen, Y.~Yang, Z.~Wu, and H.~Li, ``Lane detection in low-light
  conditions using an efficient data enhancement: Light conditions style
  transfer,'' in \emph{IEEE Intell. Veh. Symp.}, 2020, pp. 1394--1399.

\bibitem{chen2017deeplab}
L.-C. Chen, G.~Papandreou, I.~Kokkinos, K.~Murphy, and A.~L. Yuille, ``Deeplab:
  Semantic image segmentation with deep convolutional nets, atrous convolution,
  and fully connected crfs,'' \emph{IEEE Trans. Pattern Anal. Mach. Intell.},
  vol.~40, no.~4, pp. 834--848, 2017.

\bibitem{ghafoorian2018gan}
M.~Ghafoorian, C.~Nugteren, N.~Baka, O.~Booij, and M.~Hofmann, ``El-gan:
  embedding loss driven generative adversarial networks for lane detection,''
  in \emph{Eur. Conf. Comput. Vis. Worksh.}, 2018, pp. 1--17.

\bibitem{chen2019pointlanenet}
Z.~Chen, Q.~Liu, and C.~Lian, ``Pointlanenet: Efficient end-to-end cnns for
  accurate real-time lane detection,'' in \emph{IEEE Intell. Veh. Symp.}\hskip
  1em plus 0.5em minus 0.4em\relax IEEE, 2019, pp. 2563--2568.

\end{thebibliography}

\begin{IEEEbiography}
  [{\includegraphics[width=1in,height=1.25in,clip,keepaspectratio]{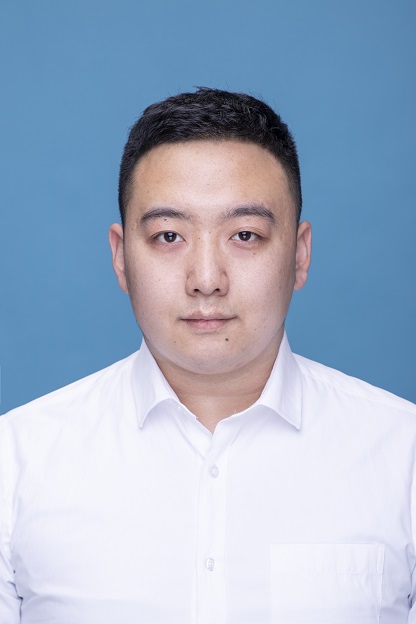}}]{Zequn Qin}
  received his master's degree in 2019 from Northwestern Polytechnical University, China. He is currently a Ph.D. candidate at Zhejiang University. His research interests include computer vision and autonomous driving.
\end{IEEEbiography}
\begin{IEEEbiography}
  [{\includegraphics[width=1in,height=1.25in,clip,keepaspectratio]{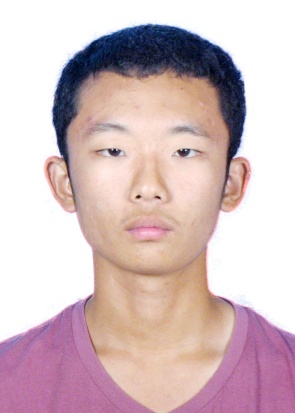}}]{Pengyi Zhang}
  received the bachelor's degree in computer science and technology from Beijing Institute of Technology, China, in 2020. He is currently pursuing the master's degree with the College of Computer Science, Zhejiang University, Hangzhou, China, under the supervision of Prof. X. Li. His current research interests are primarily in computer vision and deep learning.
\end{IEEEbiography}
\vskip 0pt plus -1fil
\begin{IEEEbiography}
  [{\includegraphics[width=1in,height=1.25in,clip,keepaspectratio]{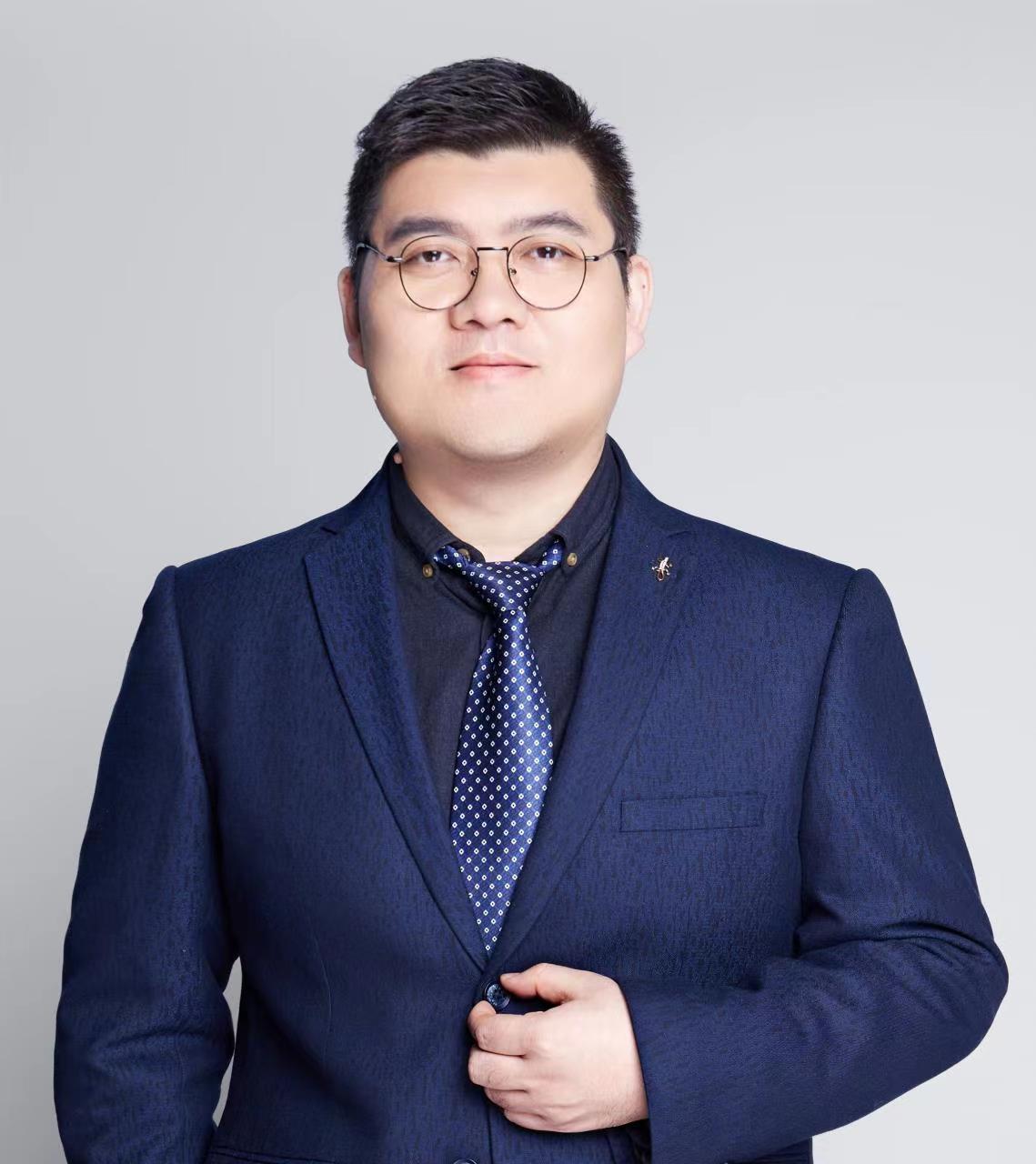}}]{Xi Li}
  received the Ph.D. degree from the Na- tional Laboratory of Pattern Recognition, Chinese Academy of Sciences, Beijing, China, in 2009. From 2009 to 2010, he was a Post-Doctoral Researcher with CNRS, Telecomd ParisTech, France. He is currently a Full Professor with Zhejiang University, China. Prior to that, he was a Senior Researcher with the University of Adelaide, Australia. His research interests include artificial intelligence, computer vision, and machine learning.
\end{IEEEbiography}

\newpage
\onecolumn
\vspace*{20pt}    
\setcounter{section}{0}
\setcounter{figure}{0}
\setcounter{table}{0}
\renewcommand\thetable{\Alph{table}}
\renewcommand\thefigure{\Alph{figure}}
\large{\textbf{\centerline{Ultra Fast Deep Lane Detection with Hybrid Anchor Driven Ordinal Classification}}}
\large{\textbf{\centerline{--Supplementary Materials}}}

\section{Overview}
These supplementary materials provide more statistics, discussions, illustrations, and tables to better cover and elaborate the motivations, advantages, and details about our work. The added topics are: statistics about the derived error band, measured FPS on CPUs and embedded chips, illustrations of the advantages of the hybrid anchor, and details about the post-processing in this work.

\section{Validating the magnified localization problem}
\begin{figure}[h]
    \centering
    \includegraphics[width=0.6\linewidth]{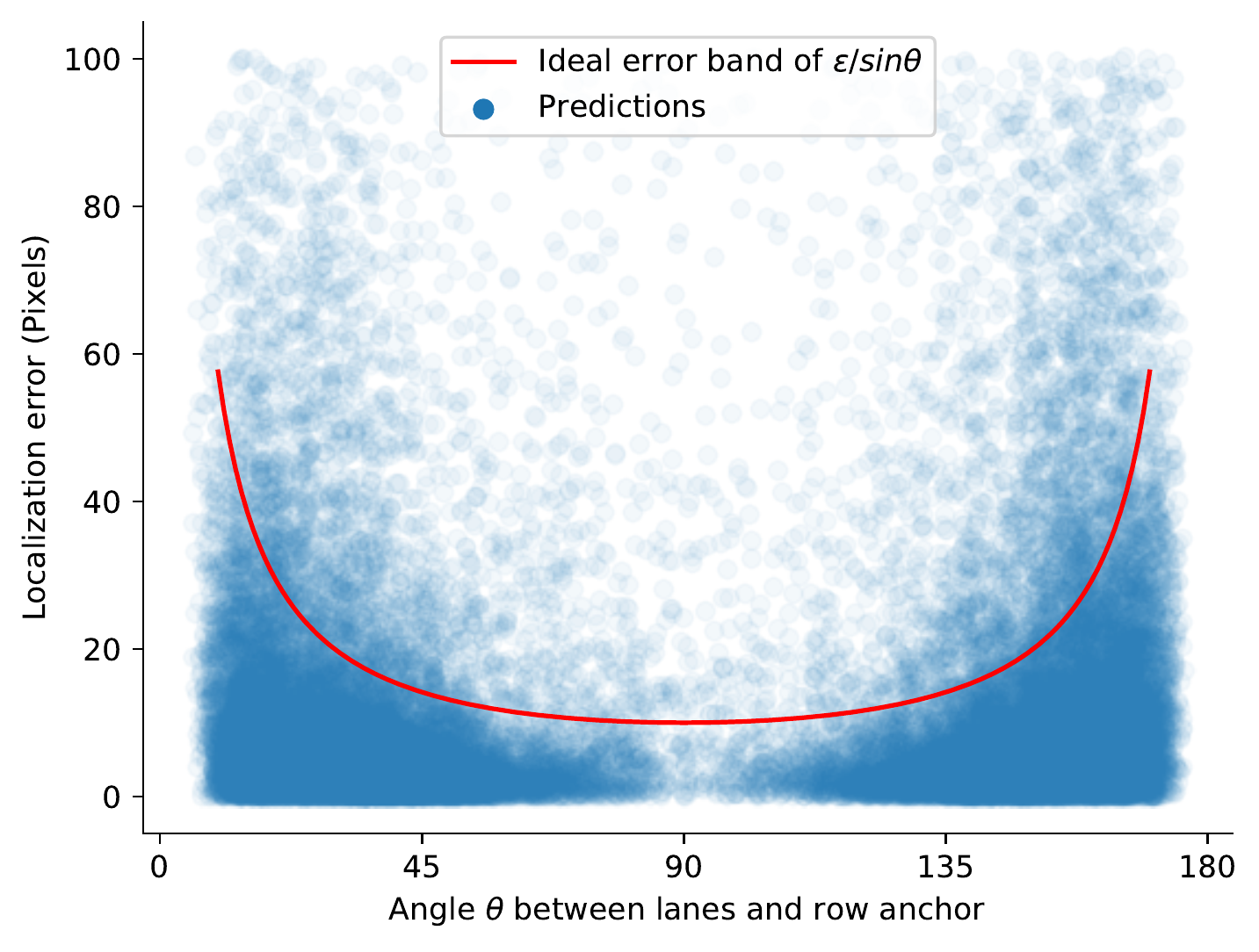}
    \caption{Illustration of the statistics of the errors vs. angles in a model with row anchors.}
    \label{fig_angle_error}
\end{figure}

To validate the magnified localization problem and the error of $\varepsilon / \text{sin} \theta$, we measure the localization errors corresponding to different angles of lanes, and the statistics are shown in \cref{fig_angle_error}. We can see that the trends of statistical results fit the mathematically derived error band of $\varepsilon / \text{sin} \theta$. This shows the magnified localization problem indeed exists.

\section{The speed on the CPUs and embedd chips}
We test our method on various CPUs and embedded chips like Nvidia TX2, as shown in \cref{tb_device,tb_device2}. We can see that our method could achieve real-time or near-real-time speed even in CPUs and embedded chips.

\begin{table}[h]
    \centering
    \caption{FPS on different devices at 320x1600 resolution with float32 precision.}
    \label{tb_device}
\begin{tabular}{ccccc}
\toprule
Device & AMD EPYC 7282 & Intel Xeon E5-2678 & Intel i9 10990K & NVIDIA Jetson TX2 \\ \midrule
FPS    & 7.7           & 7.9                & 15.0           & 36.3 \\
\bottomrule
\end{tabular}
\end{table}
\begin{table}[h]
    \centering
    \caption{FPS on different devices at 320x800 resolution with float32 precision.}
    \label{tb_device2}
\begin{tabular}{ccccc}
\toprule
Device & AMD EPYC 7282 & Intel Xeon E5-2678 & Intel i9 10990K & NVIDIA Jetson TX2 \\ \midrule
FPS    & 15.4           & 15.9                & 29.8           & 36.3 \\
\bottomrule
\end{tabular}
\end{table}

\section{Demonstrating of the advantage of hybrid anchor} 
\begin{figure}[th]
    \centering
    \includegraphics{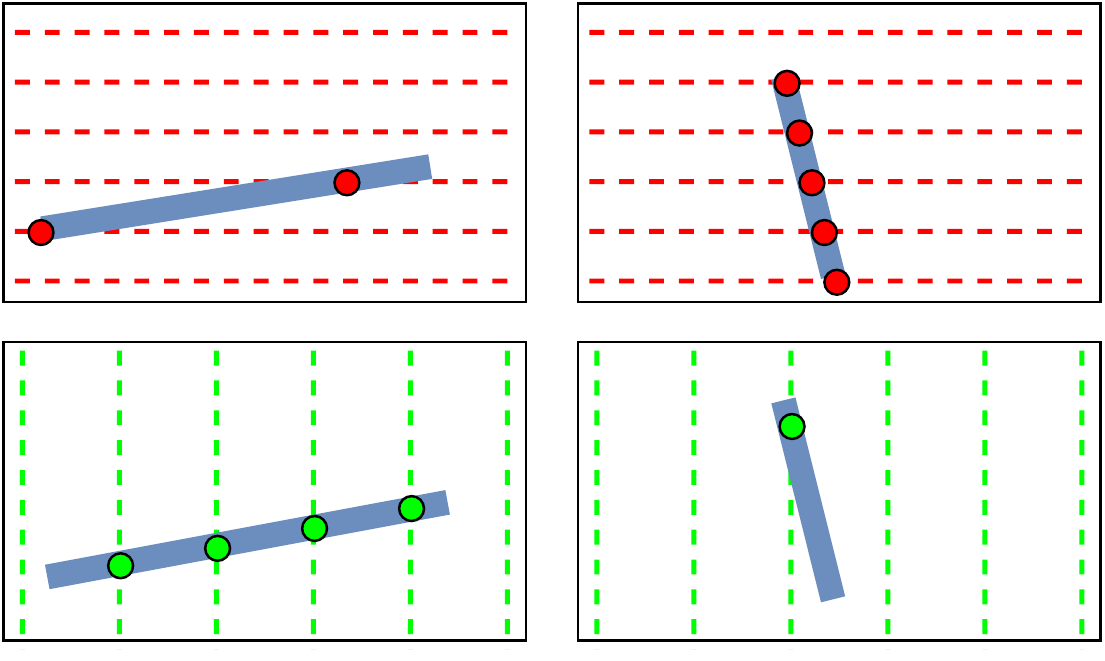}
    \caption{Illustration of the properties of different types of anchors. Red and green dashed lines represent the row and column anchors, respectively. With the same number of six anchors, row anchors could model vertical lines better while column anchors could fit horizontal lines better.}
    \label{fig_increase}
\end{figure}

To better illustrate the properties of different types of anchors, we show an example in \cref{fig_increase}. From \cref{fig_increase}, we can see that different anchor systems have different preferences in modeling lines. Specifically, row anchor is more appropriate for vertical lanes since vertical lane have large variance in vertical directions and can be more intersected by row anchors than column anchors. And this is also true for column anchors with horizontal lanes. By appropriately adopting different anchor systems for different kinds of lines, hybrid anchors could be more efficient and effective than single row or column anchors.

\section{The hidden premise of the anchor-based method}
There is a hidden premise of our method, i.e., each horizontal / vertical line should only have one intersection point with one lane instance. In this way, we discuss the effectiveness of our method when facing multiple-intersections for extremely curved or complex roads. 

\begin{figure*}[h]
    \centering
    \includegraphics[width=0.7\linewidth]{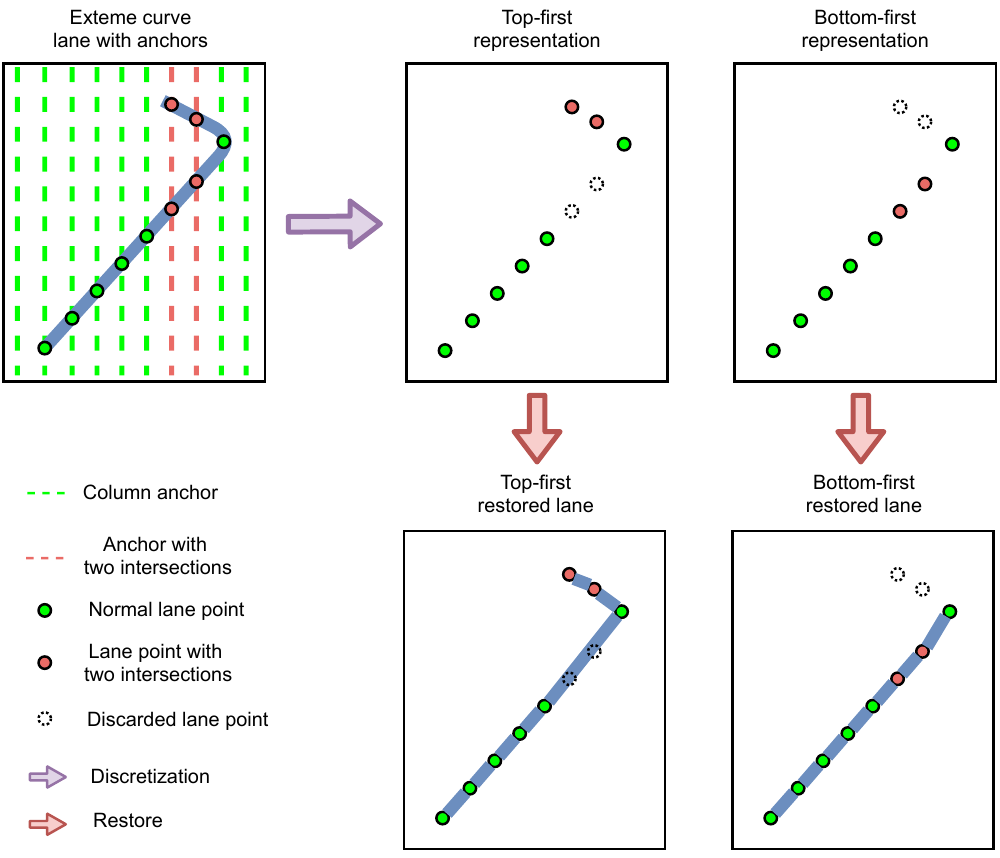}
    \caption{illustration of the multiple-intersection problem.}
    \label{fig_two_intersection}
\end{figure*}

Generally, there are two kinds of results when facing the multiple-intersection problem, which are top- and bottom-first representations. The top-first method preserves the top intersection while the bottom-first one preserves the bottom intersection. To better illustrate the problem of multiple-intersection, we show an example in \cref{fig_two_intersection}.

From \cref{fig_two_intersection} we can see that for both top- and bottom-first methods, although the restored lanes with the multiple-intersection problem are slightly inaccurate or incomplete, they are still valid. For the top-first method, the results are not much affected. The bottom-first method's results are slightly shorter than the ground truth, which is acceptable since the major lower parts of lanes are still intact. In practice, we use bottom-first representation in our work.

\section{Post-processing}
In post-processing, we would remove invalid lane predictions that are very short. For the original row-based method, we set a very loose threshold (at least two lane points) since a horizontal lane could have very few intersections with the row anchors. While for the hybrid anchor method, we can set a more aggressive threshold because hybrid anchor could represent both horizontal and vertical lanes well with denser lane points. 

This difference causes the different results of our conference version and the performance in Table 4 of the journal paper (68.4 vs. 66.09). In short, 68.4 is the row-based method with the original loose post-processing, and 66.09 is the row-based method with the aggressive post-processing of hybrid anchor.

\end{document}